\documentclass{ieeetj}
\usepackage{cite}
\usepackage{amsmath,amssymb,amsfonts}
\usepackage{algorithm}
\usepackage{algpseudocode}
\usepackage{graphicx,color}
\usepackage{textcomp}
\usepackage{xcolor}
\usepackage{hyperref}
\hypersetup{hidelinks=true}
\def\BibTeX{{\rm B\kern-.05em{\sc i\kern-.025em b}\kern-.08em
    T\kern-.1667em\lower.7ex\hbox{E}\kern-.125emX}}
\AtBeginDocument{\definecolor{tmlcncolor}{cmyk}{0.93,0.59,0.15,0.02}\definecolor{NavyBlue}{RGB}{0,86,125}}

\def\OJlogo{\vspace{-4pt}$<$Society logo(s) and publication title will appear here.$>$}
\def\seclogo{\vspace{10pt}$<$Society logo(s) and publication title will appear here.$>$}

\def\authorrefmark#1{\ensuremath{^{\textbf{#1}}}}

\DeclareUnicodeCharacter{2217}{\ast}
\begin{document}
\receiveddate{XX Month, XXXX}
\reviseddate{XX Month, XXXX}
\accepteddate{XX Month, XXXX}
\publisheddate{XX Month, XXXX}
\currentdate{XX Month, XXXX}
\doiinfo{XXXX.2022.1234567}

\markboth{}{Author {et al.}}

\title{Adaptive Dual-Mode Distillation with Incentive Schemes for Scalable, Heterogeneous Federated Learning on Non-IID Data}

\author{\uppercase{Zahid Iqbal}\authorrefmark{1}
}

\affil{Department of Computer Science, University of Gujrat, PK}

\corresp{Corresponding author: Zahid Iqbal (e-mail: zahid.iqbal@uog.edu.pk).}

\begin{abstract}
It has been increasingly difficult to effectively use the vast amounts of valuable, real-time data generated by smart devices for machine learning model training due to privacy concerns. Federated Learning (FL) has emerged as a promising decentralized learning (DL) approach that enables the use of distributed data without compromising user privacy. However, FL poses several key challenges. First, it is frequently assumed that every client can train the same machine learning models, however, not all clients are able to meet this assumption because of differences in their business needs and computational resources. Second, statistical heterogeneity (a.k.a. non-IID data) poses a major challenge in FL, which can lead to lower global model performance. Third, while addressing these challenges, there is a need for a cost-effective incentive mechanism to encourage clients to participate in FL training. In response to these challenges, we propose several methodologies: DL-SH, which facilitates efficient, privacy-preserving, and communication-efficient learning in the context of statistical heterogeneity; DL-MH, designed to manage fully heterogeneous models while tackling statistical disparities; and I-DL-MH, an incentive-based extension of DL-MH that promotes client engagement in federated learning training by providing incentives within this complex federated learning framework. Comprehensive experiments were carried out to assess the performance and scalability of the proposed approaches across a range of complex experimental settings. This involved utilizing various model architectures, such as ResNet18, DenseNet, and ResNet8, in diverse data distributions, including IID and several non-IID scenarios, as well as multiple datasets, including CIFAR10, CIFAR100, CINIC10, FMNIST, and MNIST. Empirical analysis with various SOTA approaches shows promising results. Experimental results demonstrate that the proposed approaches significantly enhance accuracy and decrease communication costs while effectively addressing statistical heterogeneity and model heterogeneity in comparison to existing state-of-the-art approaches and baselines. The experimental outcomes demonstrate significant performance gains, with DL-SH improving global model accuracy by 153\% over standard FL, and I-DL-MH achieving a 225\% improvement under non-IID conditions.
\end{abstract}

\begin{IEEEkeywords}
Federated Learning, privacy preserving machine learning, deep learning
\end{IEEEkeywords}


\maketitle

\section{Introduction}
\label{sec:introduction}
\PARstart{T}{h}{e} widespread integration of smart devices, including mobile phones, tablets, and wearables, has equipped individuals with robust computational capabilities, extensive memory storage, and advanced sensors. This shift from conventional laptops and desktops to smart devices as the primary computing platforms is evident, driven by the incorporation of AI-based smart chips, such as those introduced by \cite{Neuromation2018}. These chips aim to enhance smart devices with neural network capabilities, generating valuable data in a decentralized manner encompassing location history, images, typing patterns, medical records, and life logging data.
Despite the considerable amount of data held by smart devices, concerns regarding privacy frequently restrict individuals from disclosing their personal information for the purpose of model training. The increasing concerns regarding privacy, particularly following data breaches associated with companies such as Facebook, have led to the implementation of rigorous data protection legislation by organizations including the European Union and numerous nations \cite{Hill2021,KPMG2017,GDPR2018}. Consequently, collecting, transferring, or integrating user data without explicit consent for any purpose has become exceedingly challenging.

In the realm of distributed learning, the traditional approach involves centralizing data for model training and distributing it to separate entities. However, evolving privacy concerns and stringent regulations make it increasingly challenging to collect real-time users' private data. Nevertheless, smart devices and entities, for instance, banks and hospitals possess valuable user data that could enhance the accuracy of applications through training deep learning models. This has prompted researchers to explore a more purely decentralized learning approach to safeguard data privacy.

In this paradigm, the intuitive strategy involves locally storing users' private data and conducting necessary computations, such as model training, on the respective devices. This approach ensures the privacy of users' personal data while capitalizing on the computational resources of client devices. Collaboration among different devices becomes paramount in training more efficient deep learning models. Various collaborative learning techniques, including those proposed by \cite{Shokri2015,Konecny2016,BrendanMcMahan2017,Jeong2018,Iqbal2021, Corinzia2019} , have been explored. Notably, Google introduced a promising decentralized learning technique known as Federated Learning (FL), attracting considerable attention within the machine learning research community. This technique involves moving the code to data (model) instead of transferring data to code (computing).

FL assumes collaboration among clients to train a global model for specific tasks. All devices collaborate through a centralized server (aggregation server), wherein the global model is distributed to active devices, trained on private data, and subsequently updated. This iterative process continues until the global model converges. The advantages of FL over traditional distributed machine learning approaches, including privacy, low latency, and efficient utilization of computational resources and network bandwidth, have been underscored.
While FL offers a decentralized framework to leverage distributed and non-IID private data from smart devices, it introduces unique challenges related to data distribution, model architecture, communication, and privacy. The decentralized and privacy-centric nature of data, coupled with communication costs and the need for personalized models, necessitates careful consideration. This paper primarily focuses on addressing model heterogeneity and statistical heterogeneity (non-IID) while minimizing communication and computation overhead. 
The central goal of federated learning is to train a single global model, assuming reasonable computational resources across all clients and the capability of all participating devices to train the same complex model architecture. However, this assumption proves unrealistic in real-world scenarios where devices with valuable data may differ in computational resources and business needs. Model heterogeneity becomes pronounced in cases where devices have varying sizes of deep networks or entirely different network architectures. The heterogeneity challenge is exacerbated when relaxing the assumption that heterogeneous clients should have the same number of target nodes in the output layer. This scenario is particularly relevant in applications where clients exhibit non-IID data. For instance, in healthcare applications, devices may require personalized models based on their categories of data, leading to unnecessary computation and communication overhead when assuming uniform output layers among collaborating models.

Recent research efforts have partially addressed the model heterogeneity challenge, although many existing approaches are constrained by strong assumptions or lack practical feasibility in FL scenarios. Some researchers have leveraged distributed multitask learning, transfer learning, and meta-learning to address model heterogeneity. However, certain approaches are limited to convex problems or exhibit scalability challenges in large FL scenarios. Model personalization techniques, such as retraining the collaboratively trained global model on users' local private data, and approaches using transfer learning and meta-learning, have also been proposed.
Notably, researchers have explored the application of distillation, a communication-efficient approach, to partially address heterogeneous models in FL settings. Distillation efficiently transfers knowledge from a trained model to an untrained model, aligning with the information exchange required in FL. However, distillation's limitations, such as its applicability to IID data distribution, pose challenges in FL settings characterized by non-IID data. A recent proposal, Federated Distillation, represents a variation of FL applied in a decentralized environment. While demonstrating the communication efficiency of codistillation compared to standard FL, the approach requires making all data distributions IID using a Generative Adversarial Network (GAN) approach.
Furthermore, in FL, when a centralized server performs aggregation on the client’s update then the server assigns some weights to these updates before aggregation. These weights are assigned based on the number of training samples on which a model was trained. However, it might not be an efficient way to assign weightage. For instance, a device might have more samples but its model is not well trained due to diff. factors including insufficient training resources, data quality etc. Similarly, a model trained on a few training examples might be more efficient. Therefore, there is a need to devise a more efficient way to assign weightage to different models’ output. For instance, it could be a more promising solution if weights are based on the client model’s learning so if a client is more confident about its output, then the server may give it more weightage.

Typically, the main objective of federated learning is to train the global model where all participating devices collaborate to share their knowledge with the global model. In federated learning settings, typically, it is assumed that clients would be voluntarily convinced to participate in federated learning training. However, in practical scenarios, it could be very difficult to convince the clients to allow someone to consume their computational and communicational resources along with valuable data with no incentives. In pioneer work \cite{BrendanMcMahan2017} of federated learning by Google, they use the complete model sharing approach where in each round, participating clients can also get the updated trained model from the server (as incentives) so they can also use that updated model on their devices. Moreover, Google has access to (android) operating systems (OS) of millions of devices, so they can return the incentives in the form of OS updates. However, for small third-party developers/organizations, it might not be possible to give incentives to participating clients in the same manner. Particularly, under fully model heterogeneous settings, this could be more difficult where all clients may have entirely different model architectures including different targets. Thus, the client cannot distill the knowledge from the global model straightforwardly. 
Recent research has explored incentive mechanisms to improve client participation in federated learning, including dynamic screening for cross-silo settings \cite{zhang2025incentive}, market-style reward allocation for inclusivity \cite{chai2025incentivizing}, and reputation-driven adaptive incentives \cite{rashid2024trustworthy}. Other studies have introduced trust-aware models that adjust rewards based on heterogeneous client behaviorsv\cite{xu2024trust} and auction-based methods to balance participation incentives with energy efficiency \cite{zhou2023incentive}. While these approaches provide important advances in fairness, trust, and cost-efficiency, most of them generally assume clients operate under homogeneous global architectures and rely on monetary, token, or selection-based rewards and have limitations in explicitly addressing the challenge of fully heterogeneous model architectures, where clients employ structurally distinct models and participation is rewarded by delivering global model updates as a utility. Thus, cost-efficient incentive schemes are required to motivate the client devices to participate in federated learning training.

In summary, in decentralized settings where a centralized server is absent to manage data distribution, clients often possess highly unbalanced and non-IID data, introducing statistical challenges. Existing techniques, while addressing these challenges, often rely on unrealistic assumptions or face implementation difficulties in real Federated Learning (FL) scenarios. Moreover, model heterogeneity, characterized by different architectures among clients, presents additional complexities that generic FL approaches struggle to handle. Furthermore, the assignment of weights to client updates based solely on the number of training samples may not be efficient. Moreover, convincing clients to participate in FL, especially in fully model-heterogeneous settings, poses challenges in terms of providing incentives. This led us to the following research questions.
\begin{enumerate}
    \item How can statistical heterogeneity be efficiently managed in FL settings using unlabeled public data?
    \item In the context of fully heterogeneous models with different architectures and target labels, how can such diversity be effectively leveraged within the current FL framework?
    \item What strategies can be employed to encourage client participation in current FL training, particularly in scenarios with fully heterogeneous models?  
\end{enumerate}
This research aims to propose a more robust and communication-efficient decentralized learning framework that appropriately addresses these research questions, focusing on learning from fully heterogeneous models while reasonably satisfying statistical heterogeneity.

\subsection{Contribution}
The contribution of this research is briefly discussed as follows:
\begin{enumerate}
    \item A confidence matrix is computed for each client by training a binary classifier on unlabeled data to address the statistical heterogeneity problem. It gives around 153\% performance gain to the global model as compared to standard FL under the most complex data distribution (non-IID) with FMNIST using ResNet18. 
    \item Cost-effective mapping and masking schemes are applied to clients' outputs to enable fully heterogeneous model architectures to participate in federated learning training under non-IID settings. It reduces the communication cost by around 99\% as compared to standard FL under the most complex data distribution (non-IID) with CIFAR100 using ResNet18.
    \item Very appealing incentives are provided to clients in the form of updated knowledge from the global model with negligible additional communication costs. Client models get almost similar performance to the global model by following this incentive approach. 
    \item Only a single round of communication/distillation is necessary, unlike the multiple rounds required by current approaches.
    \item Comprehensive evaluations were conducted on various experimental settings, including
    \begin{itemize}
        \item Various homogeneous and heterogeneous model architectures, including ResNet 18, ResNet 8, DenseNet
        \item Various datasets, including MNIST, FMNIST, CIFAR10, CIFAR100, CINIC10
        \item Various data distributions from a simple IID scenario to multiple non-IID scenarios (of varying complexity level)
        \item Multiple client settings (2, 5, 10)
    \end{itemize}

\end{enumerate}


\section{Background and Literature Review}
\label{sec:literature-review}
\begin{figure}[H]
    \centering
    \includegraphics[width=1\linewidth]{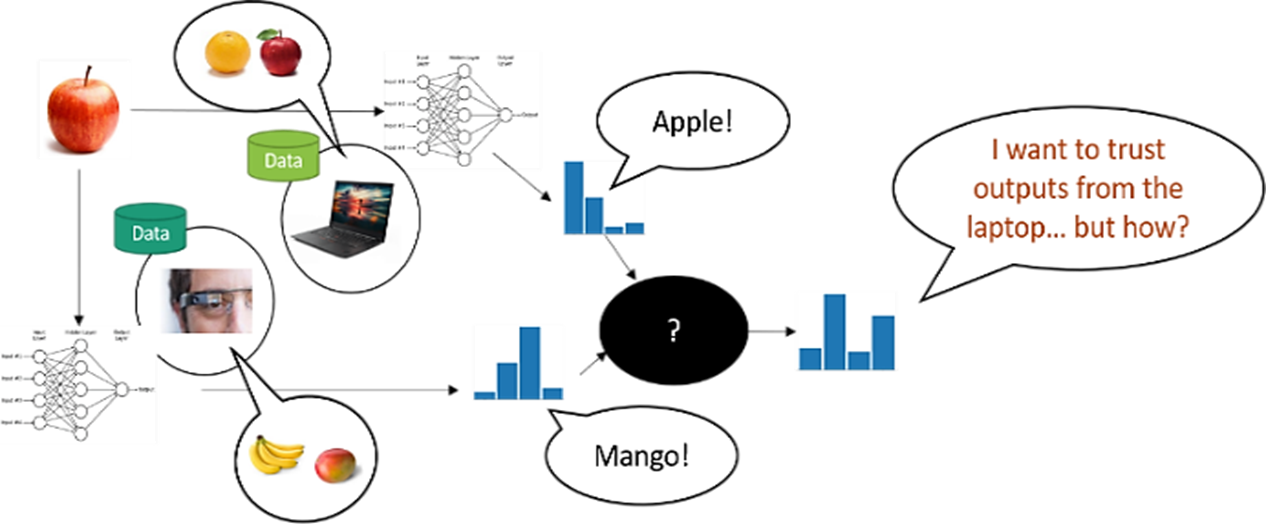}
    \caption{An example of non-IID data}
    \label{fig:statistical-hetero-problem}
\end{figure}

In decentralized settings we typically assume there is no centralized server to properly manage the distribution of data. Thus, it is very likely that clients would have highly unbalanced and non-IID data, as each device or user may have distinct preferences. FedAvg \cite{BrendanMcMahan2017}, a state-of-the-art algorithm based on SGD, shows that it can handle a certain amount of non-IID data; however, this study \cite{Zhao2018} has shown empirically that for highly skewed non-IID data, the performance of the convolutional neural network, trained using FedAvg can drop reasonably by 51\% on CIFAR10, 11\% on MNIST and 55\% for keyword spotting datasets. Therefore, FedAvg does not effectively handle the skewed non-IID data which is natural and expected data distribution in the federated learning setting.Figure \ref{fig:statistical-hetero-problem} illustrates an example of non-IID data distribution.

Some researchers proposed the idea to share some of the private data of devices or share some proxy labelled data to handle the statistical heterogeneity \cite{Jeong2018,Zhao2018,Huang2020} to make the data distributions of devices as IID. Like, Zhao et al. \cite{Zhao2018} show that accuracy reduction, in the case of non-IID data, could be attributed to weight divergence when two different training processes having the same weight initialization get different weights. They propose that if we can leverage the globally shared data (having uniform distribution over all classes) by distributing it to all clients, then it can reduce the weight divergence between distribution on the different devices. This weight divergence could be quantified using EMD (Earth Moving Distance), and in return, it would increase the accuracy of the model. Their approach seems infeasible, as arranging and communicating uniform distribution of data over all classes could be challenging and can create overhead for communication. 
Jeong et al. \cite{Jeong2018} proposed FAug (Federated Augmentation), which uses the concept of conditional GAN (Generative Adversarial Network) to produce the missing label samples on client devices by data augmentation. Where each client is required to identify and upload the missing target labels, in its distribution, to the server. The server oversamples these target labels to train the conditional GAN. Finally, all devices download this trained GAN to produce missing target labels in their distribution. As the target labels of each device may reveal some private and sensitive information with the server or with other devices (which have GAN trained on all devices’ data and that could be used to infer the other’s target labels), it requires all devices to additionally upload the redundant samples, other than target labels, on the server to handle the privacy issue at the cost of extra communication overhead. However, this method works with the assumption that client devices would agree to share their private data with the server. This seems an almost impractical solution and violates the key idea of FL, i.e., privacy.
Huang et al. \cite{Huang2020} proposed an effective approach to address the statistical heterogeneity challenge. Participating clients compare the cross-entropy loss with the median loss of the last round, and if the current loss is higher than the previous loss, then client models are retrained to further decrease the loss before aggregation at the server end. However, they also assume that there should be some publicly labelled (annotated) data. That again cannot be confirmed in a real-life scenario where some data, say healthcare data, might not be available in the annotated form, and annotating the data may require a lot of prudent efforts.

Some researchers \cite{Wang2016,Smith2017,Corinzia2019} have shown that the natural way to address the statistical challenge (non-IID) of data is Multitask Learning (MTL), where the goal is to learn from each node, having separate but related models, simultaneously. Here, each node represents a task that possesses its private data, and the goal is to learn from these related but different tasks. For instance, Smith et al. \cite{Smith2017} use the MTL in the federated learning setting. In MTL, an additional term is included in the loss function to model the relationship among tasks. They use the correlation matrix to measure the client similarity and train separate but related models for each device (task) using a shared representation on the server. However, their method only works for convex optimization problems and is not scalable to a large population. Furthermore, Lim et al. \cite{Lim2019} argue that this approach is not very suitable in federated learning scenarios when a specific task (model) doesn’t possess its local data or may have very few training samples. Corinzia et al. \cite{Corinzia2019} employ the concept of a Bayesian network to connect all clients with the server and perform variational inference during learning. Their method can properly handle the non-convex problem; however, it is much more costly to scale it to a vast federated network, as it refines the client models sequentially.

Duan et al. \cite{Duan2019} revealed that model performance could also deteriorate due to global imbalance (when local distributions of data across all clients have class imbalance). It first removes the global imbalance by data augmentation, where all devices first share their data distribution with the centralized server. Then, before performing local model training, each device first performs data augmentation on imbalanced classes to make a balanced distribution. Subsequently, it employs the concept of mediators to combine training samples of relevant devices (selection is performed by calculating KL divergence between local and uniform distribution) based on their distributions to make it a uniform distribution. So, finally, this combined training (model) is shared with the global server for federated aggregation.

Some researchers \cite{Yurochkin2019,Li2020,Wang2020,Xiao2020,Liu2021} have identified the limitations of the standard FedAvg algorithm, particularly when clients have statistical heterogeneity. Lim et al. \cite{Lim2019} question the performance of the standard FedAvg algorithm and suggest that standard aggregation is probably not the best aggregation way. By using the Mutual Information (MI) and different distance metrics, they demonstrate that with the increase in the number of iterations, correlation (MI) increases; however, in parallel, the distance of parameters also increases. Similarly, Xiao et al. \cite{Xiao2020} mention the three limitations of FedAvg, i.e., 1) it cannot be applied to non-differentiable methods, 2) it usually requires many communication rounds, and 3) it is primarily designed for the cross-device setting. While addressing these limitations, they propose FedKT for cross-silo scenarios, which can learn from both differentiable and non-differentiable models. Moreover, Li et al. \cite{Li2020} explain that due to the permutation invariance of NN, simple model parameter aggregation (FedAvg) may have a very negative impact, so they propose PFNM, a probabilistic Federated Neural Matching algorithm that performs the matching among clients’ NN neurons before averaging them. Yurochkin et al. \cite{Yurochkin2019} further extend this approach and propose a layer-wise matching approach (FedMA) and apply this approach to modern CNNs and LSTMs. Wang et al. \cite{Wang2020} try to reduce Aggregation Error (AE) by constructing a definitely convex global posterior using a Gaussian product method to obtain the global expectation and co-variance by multiplying local posteriors. On the client side, they proposed a new Federated Online Laplace Approximation (FOLA) method to obtain online local posterior probabilistic parameters which can directly be leveraged in the FL framework.

Recently, some researchers \cite{Luo2021} have proposed a very effective approach to address the non-IID challenge. They argued that non-IID mainly adds bias to the classifier of any trained model. They argued that applying some calibration to the classifier can greatly improve the performance of any existing global model architecture. Another work ODIN \cite{Liang2018} by Facebook has shown that it can effectively address the out-of-distribution (non-IID) problem. They try to separate the softmax distribution score between in-distribution and out-of-distribution images by adding a small perturbation along with temperature scaling to input images. They demonstrated that ODIN could reduce the False Positive Rate (FPR) from 34.7\% to 4.3\% on the CIFAR10 dataset while the True Positive Rate (TPR) was 95\%. Based on its effectiveness, this work also considers it as a benchmark for proposed approaches.

\begin{figure}
    \centering
    \includegraphics[width=1\linewidth]{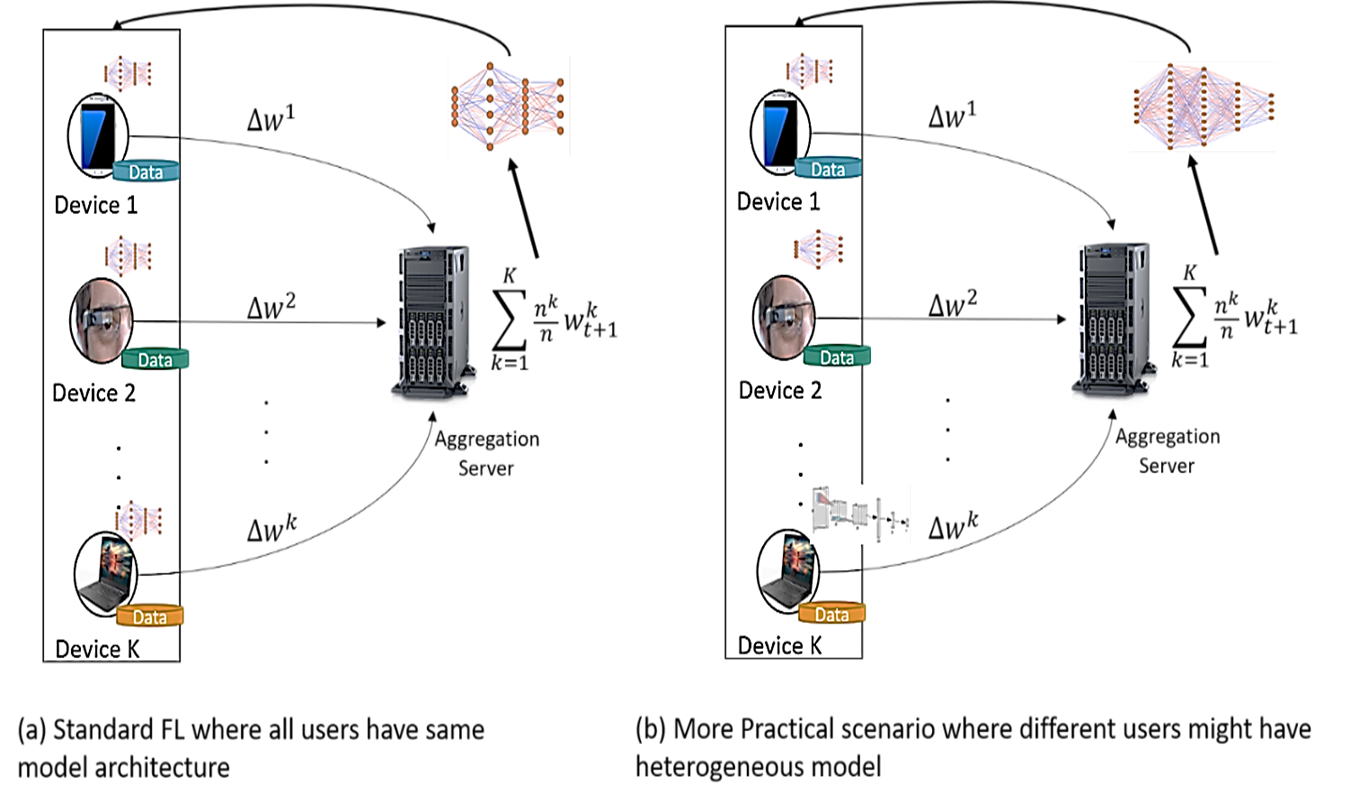}
    \caption{Difference between (a) standard federated learning, where all devices are required to have the same model architecture and (b) a More practical scenario of federated learning, where different users might have a different model architect}
    \label{fig:model-hetero-problem}
\end{figure}

Typically, system heterogeneity is defined as where devices possess varied computational resources like different memory, processors, battery limits, active times, etc. More specifically, in model heterogeneity cases, due to varied computational resources and different business requirements, it is intuitive that devices may have varying sizes of deep networks (different no. of layers) or may have completely different network architectures. For instance, some devices may be using CNN, some devices may be using ResNet, while some devices may opt for the Inception network. Figure \ref{fig:model-hetero-problem} shows the model heterogeneity problem.

Having the same model architecture would not only overburden the communication (already facing high communication challenges in federated learning) but would also increase the computation complexity for devices where low-resourced devices may result in the form of stragglers or stagnant data. Probably, some devices might possess immensely valuable data however unable to train the same complex model. Therefore, it is intuitive that models should possess the proper number of nodes in their output layer to avoid unnecessary computation and communication overhead.

We can divide the model heterogeneity into two general categories. 1) Where different models need to be personalized based on different geographical or personal preferences, like in the case of next word prediction, for the sentence “I love to visit …”, there would be customised predictions for different users living in separate geographical locations or with distinct preferences. 2) where various models might have diverse architecture due to varied computational resources or different business needs.

\begin{figure*}[!htbp]
    \centering
    \includegraphics[width=\linewidth]{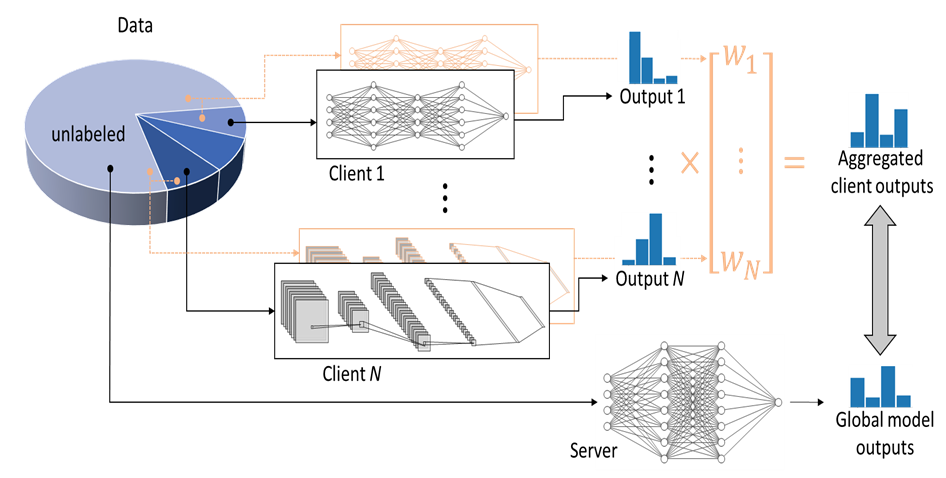}
    \caption{Decentralized Learning with Statistical Heterogeneity (DL-SH)}
    \label{fig:DLSH}
\end{figure*}

The primary goal of federated learning is to train a unique global model. Standard FL works with the assumption that all clients would possess reasonable computational and communicational resources to train the same model architecture. Due to such a primary assumption of FL, most of the work has been performed with the same assumption of homogeneous models, and most of the work has been performed for category 1 of model heterogeneity (model personalisation based on personal preferences or distinct geographical locations), and very few works have addressed this potential problem of category 2 of model heterogeneity (clients contain diverse model architecture), and in a significantly limited fashion.

To make a global model personalized, most personalization techniques suggest retraining the (collaboratively trained) global model on the users’ local private data \cite{DBLP:conf/interspeech/SimZB19}. Some researchers have proposed model personalization approaches using transfer learning \cite{Wang2019,Schneider2019MassPO,Mansour2020ThreeAF}. In transfer learning, usually, the last layers of a trained model are replaced with new layers to leverage the learnt knowledge of the trained model on some new tasks. Some researchers suggest freezing the initial layers of a trained global model and retraining only the last few layers on the local private data of individual clients.

Some researchers \cite{Finn2017ModelAgnosticMF,Jiang2019,Khodak2019AdaptiveGM,Fallah2020PersonalizedFL} have also leveraged meta-learning to solve the personalization problem. Multitask learning \cite{Caruana1998} has also been widely used by different researchers \cite{Wang2016, Liu2017,Ruder2017,Smith2017,Corinzia2019,Sattler2019} to address the model personalization challenge. They leverage distributed multitask learning to train separate but related models. They try to train a personalized model for each distribution (as in FL, we assume the non-IID distribution). In addition to limitations of scalability and feasibility, these approaches address the model heterogeneity scenarios in a very limited fashion.

Some researchers \cite{Jeong2018, Li2019, Malinin2019, Vongkulbhisal2019, He2020a, Itahara2020, Lin2020,Mansour2020ThreeAF,Mirzadeh2019} have leveraged the knowledge distillation to allow clients to use customized local models having a different number of layers. Therefore, it allows the clients to use diverse model architecture while collaborating in federated learning. However, most of them only partially address this problem \cite{Jeong2018, He2020a, Itahara2020, Lin2020, Ma2020} or some have proposed their approaches for a different domain \cite{Banitalebi-Dehkordi2021, Li2021a} where the authors \cite{Banitalebi-Dehkordi2021} argue that the distillation approach is different for object detection from the object classification problem.

\section{Proposed Methodology}
\label{sec:methodology}
\subsection{Effectively Handling Statistical Heterogeneity In Decentralized Learning}
This paper first proposes a very effective approach named Decentralized Learning with Statistical Heterogeneity (DL-SH) that leverages the unlabelled public data using distillation \cite{Hinton2015} to effectively address the statistical heterogeneity challenge. This approach also employs the concept of one-round learning to further address the client dropout and stagnant data problems. Where, due to limited computational and communicational resources, clients might not be able to participate in each training round, resulting in low global model performance. By using one-round communication, DL-SH also greatly reduces the computation and communication cost, which is also a great challenge for devices having limited resources.

As illustrated in Figure \ref{fig:DLSH}, each client \(M_i\)  has its own private data \(D_i=(X_i,Y_i )\). Each client is asked to train a binary classifier \(C\)  which also has the same model architecture as the client model. Only the last layer of the client model is replaced with a 2-node output layer. The primary purpose of this binary classifier \(C\) is to distinguish the client training data \(X\) from unlabelled public data \(X_{\text{dist}}\).
More specifically, it tries to learn whether a specific sample \( x \in X_{\text{dist}} \)  belongs to the training data \(X\) of a client. If the sample  \(x\) belongs to the client training data, \(X\) then the global model will give more weightage to this client against the sample \(x\) while training on \( X_{\text{dist}} \). Thus, it calculates a confidence matrix \(w\) to measure the confidence value of each \( x \in X_{\text{dist}} \) for each client. Each client sends its logits \( \vartheta \) and confidence matrix \(w\) to the server, where the server takes the weighted aggregation of clients’ knowledge by\( Y_g \leftarrow \prod(\vartheta, w) \). Finally, the global model distils the knowledge from clients by training on \( X_{\text{dist}} \) using \(Y_g\)  as the target labels. DL-SH gives around 153\% performance gain to the global model as compared to standard FL and around 150\% improvement as compared to the ODIN approach against most complex data distribution settings (non-IID) in the FMNIST dataset. 
\begin{algorithm}[H]
\caption{DL-SH (CLIENT-TRAINING)}
\label{alg:dlsh_client}
\textbf{Description:} $M$ clients are indexed by $M_i$. $E$ is the number of local epochs for each client $M_i$. $C_i$ represents the binary classifier of client $i$. $E_{\text{embed}}$ is the number of epochs for training each binary classifier $C_i$. $M_{\text{tgt}}$ is the global model on the server, trained for $E_g$ epochs. $\eta$ is the learning rate. $D_i = (X_i, Y_i)$ is the local training data for client $M_i$. $D_{\text{dist}} = X_{\text{dist}}$ is the public unlabeled data for distillation. $F(\vartheta, \text{input})$ denotes the prediction function.

\begin{algorithmic}[1]
\For{each client $i \in M$}
    \State $B \gets$ split $D_i$ into batches of size $B$
    \For{each local epoch $e = 1$ to $E$}
        \For{each $b \in B$}
            \State $(y, x) \gets$ \Call{get\_train\_samples}{$b$}
            \State $\vartheta \gets \vartheta - \eta \nabla_{\vartheta} \{ \phi(y, F(\vartheta, x)) \}$
        \EndFor
    \EndFor

    \Comment{Train binary (embed) classifier $C_i$ after local model training}
    \State $X_{\text{embed}} \gets \text{concatenate}(X_i, X_{\text{dist}})$
    \State $Y_1 \gets \text{zeros}(\text{len}[X_i])$
    \State $Y_2 \gets \text{ones}(\text{len}[X_{\text{dist}}])$
    \State $Y_{\text{embed}} \gets \text{concatenate}(Y_1, Y_2)$
    \State $D_{\text{embed}} \gets (X_{\text{embed}}, Y_{\text{embed}})$

    \State $B \gets$ split $D_{\text{embed}}$ into batches of size $B$
    \For{each local epoch $e = 1$ to $E_{\text{embed}}$}
        \For{each $b \in B$}
            \State $(y, x) \gets$ \Call{get\_train\_samples}{$b$}
            \State $w \gets w - \eta \nabla_{w} \{ \phi(y, F(w, x)) \}$
        \EndFor
    \EndFor
\EndFor
\end{algorithmic}
\end{algorithm}

Algorithm \ref{alg:dlsh_client} represents the working of clients’ training in DL-SH. In lines 1-8, all clients train their local models on their own private data \(D_i=(X_i,Y_i )\) . All clients train their model for E number of epochs and return the logits \( \vartheta \). Contrary to standard distillation approaches, the DL-SH adaptively aggregates the client models, and then these adaptively calculated outputs are used to train the global model. More specifically, rather than performing simple aggregation on client models’ output, the proposed approach introduces the concept of a confidence matrix w, which basically informs how much a client is confident about its predictions against provided classes.  Therefore, the new objective of the proposed problem becomes as follows:

\begin{equation}
L(D_{\text{dist}}) = \left[ l_1 \left( \sum_i w_i(x) M_i(x), M_{\text{tgt}}(x) \right) \right]
\end{equation}
Here \( l_{\text{i}} \) is categorical cross-entropy loss, and \( w_{\text{i}} \) (x) is the confidence matrix. 
To calculate the confidence matrix \( w_{\text{i}} \) (x) with unlabeled data, each client is asked to train a binary classifier \( C_i(x) = [0, 1] \) with sigmoid outputs. The objective of this classifier is to distinguish the client’s local private data  \( X_i \in D_i \quad \text{and} \quad X_{\text{dist}} \in D_{\text{dist}} \)
. Each client’s binary classifier is created by modifying the client's local model. More specifically, only the last layer of the client's local model is replaced by a 2-node dense layer. It is assumed that each client i trained its binary classifier \( C_{\text{i}} \) optimally; thus, the prediction of \( C_{\text{i}} \)(x) can be expressed as follows:
\begin{equation}
C_i^*(x) = \frac{p_i(x)}{p_{\text{dist}}(x) + p_i(x)}
\end{equation}
Here \( p_{\text{i}} \)(x) is the probability of sample (x) on \( x_{\text{i}} \) and \( p_{\text{dist}} \)(x) is the probability of sample (x) on \( x_{\text{dist}} \). \( C_i^{*} \) can also be represented as
\begin{equation}
C_i^* = 1-\frac{p_{\text{dist}}(x)}{p_{\text{dist}}{\text{(x)}} + p_i(x)}
\end{equation}
By fixing the value of x and considering \( p_{\text{dist}} \)(x) as a positive constant,\( C_i^{*} \)
 (x) monotonically increases with \( p_{\text{i}} \) (x) within \( p_i(x) \in [0, 1] \)
. It implies that it \( C_i^{*} \)
 would compute higher values when sample  x is more likely to be present in \( x_{\text{i}} \) . Thus, the client \( M_{\text{i}} \) would be more confident about its prediction \( M_{\text{i}} \) (x).
More technically, after completing the training of local models on their private data, each client trains its own binary classifier \( C_{\text{i}} \)  on the client’s private data \( X_{\text{i}} \) and public data \( X_{\text{dist}} \).  To train the binary classifier, in lines 9-13, each client first combines the client's private training samples along with the public unlabeled training samples by

\( X_{\text{embed}} \leftarrow \text{concatenate} \)
(\( X_{\text{i}} \)  ,\( X_{\text{dist}} \) )

For target variables, training data is assigned as label 0, and public unlabeled data is assigned as 1. Finally, these target labels are combined as well:

\( Y_{\text{1}} \leftarrow \text{zeros}(len[X_i ]) \)

\( Y_{\text{2}} \leftarrow \text{ones}(len[X_\text{dist}) ] \)

\( Y_{\text{embed}}\leftarrow \text{concatenate} \)
(\( Y_{\text{1}} \)  ,\( Y_{\text{2}} \) )

\begin{figure*}[!htbp]
    \centering
    \includegraphics[width=\linewidth]{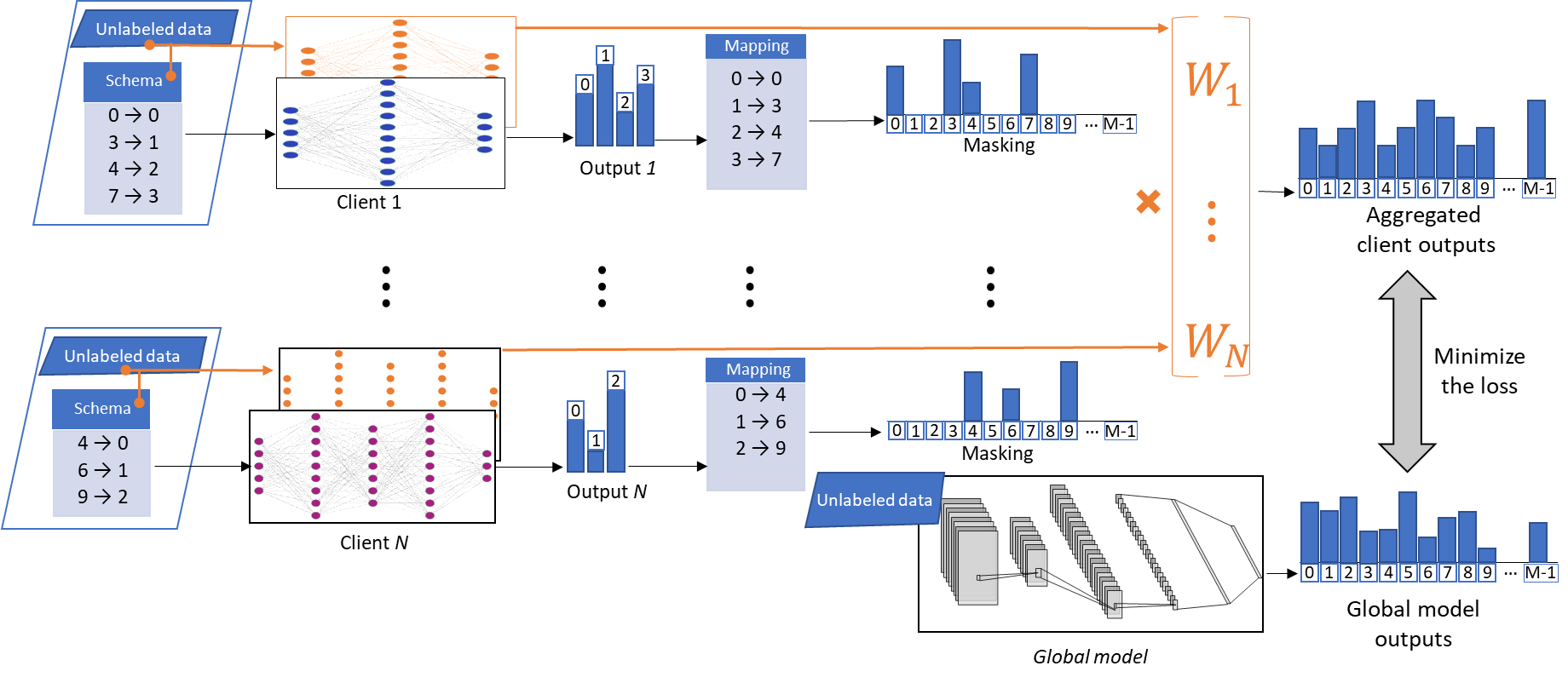}
    \caption{Decentralized learning with model heterogeneity (DL-MH))}
    \label{fig:3.8DLMH}
\end{figure*}

So, now, there is a new dataset for binary classifier training i.e.
\( D_{\text{embed}} \leftarrow \text{concatenate} ( X_{\text{embed}},Y_{\text{embed}}\)). In lines 15-20, the binary classifier \( C_{\text{i}} \) performs the training on \( D_{\text{embed}} \leftarrow \text{concatenate} (X_{\text{embed}},Y_{\text{embed}}\)) for \( E_{\text{embed}} \)  a number of epochs. After training, the binary classifier \( C_{\text{i}} \) returns the confidence matrix w.

\begin{algorithm}[H]
\caption{DL-SH (SERVER TRAINING)}
\label{alg:dlsh_server}
\textbf{Description:} $M$ clients are indexed by $M_i$. $E$ is the number of local epochs for each client $M_i$. $C_i$ represents the binary classifier of client $i$. $E_{\text{embed}}$ is the number of epochs for training each binary classifier $C_i$. $M_{\text{tgt}}$ is the global model on the server, trained for $E_g$ epochs. $\eta$ is the learning rate. $D_i = (X_i, Y_i)$ is the local training data for client $M_i$. $D_{\text{dist}} = X_{\text{dist}}$ is the public unlabeled data for distillation. $F(\vartheta, \text{input})$ denotes the prediction function. The global model $M_{\text{tgt}}$ performs post-distillation training on $D_g = (X_{\text{dist}}, \text{logits})$, \text{get\_train\_samples(x)} simply returns training data.

\begin{algorithmic}[1]
\For{each client $i \in M$}
    \State $w_i \gets \frac{\exp(w_i / T)}{\sum_j \exp(w_j / T)}$
\EndFor

\State $\vartheta \gets \sum_{i=1}^M \vartheta_i$
\State $w \gets \sum_{i=1}^M w_i$
\State $Y_g \gets \prod(\vartheta, w)$
\State $D_g \gets (X_{\text{dist}}, Y_g)$

\State $B \gets$ split $D_g$ into batches of size $B$
\For{each global epoch $e = 1$ to $E_g$}
    \For{each $b \in B$}
        \State $(y, x) \gets$ \Call{get\_train\_samples}{$b$}
        \State $w \gets w - \eta \nabla_w \{ \phi(y, F(w, x)) \}$
    \EndFor
\EndFor
\end{algorithmic}
\end{algorithm}

After completing the client’s training, the server starts its distillation and training process as indicated in Algorithm \ref{alg:dlsh_server}. In line 2, the server applies the softmax(w).   When a client model \( M_{\text{i}} \) is trained on non-IID data and provides a variety of responses to \( x \in X_{\text{dist}} \), then confidence \( w_{\text{tgt}} \) (x) should be higher for that client who has observed more similar samples of the same classes in their private training data. Thus, this client would be more confident to inform the global model about the output of sample x. The confidence matrix \( w_{\text{i}} \) (x) is calculated by applying the softmax activation function on \( C_{\text{i}} \) (x) as follows:

\begin{equation}
W_i(x) = \frac{\exp\left(C_i\left(\frac{x}{T}\right)\right)}{\sum_j \exp\left(C_j\left(\frac{x}{T}\right)\right)}
\label{eq:3.4}
\end{equation}

Here the purpose of applying the softmax is to ensure that \( \sum_i w_i(x) = 1 \)
. Moreover, T is the temperature parameter to control the smoothness of the output. 
After applying softmax, the server performs the aggregation on clients’ logits \( \vartheta \) and confidence matrix w in lines 4 and 5, respectively. To perform aggregation, the server takes the average of all clients’ logits \( \vartheta \) and similarly takes the average of all clients’ confidence matrices w. After that, the server multiplies the clients’ logits \( \vartheta \)  with their confidence matrix w  to compute target labels \( Y_{\text{g}} \). Then, the global model \( M_{\text{tgt}} \)  used this \( Y_{\text{g}} \) along with public unlabeled data \( X_{\text{dist}} \) as follows:

\( D_g \leftarrow (X_{\text{dist}}, Y_g) \)

Finally, the server trains its global model \( M_{\text{tgt}} \)  on \( D_{\text{g}} \) for E number of epochs.
\subsection{A Robust Decentralized Learning Approach To Leverage Heterogeneous Clients With Non-iid Data}
To address the model heterogeneity challenge in federated learning while satisfying the statistical heterogeneity, this paper proposed a second approach named Decentralized Learning with Model Heterogeneity (DL-MH). Most of the work in federated learning follows the standard assumption that all participating clients would have enough computational and communicational resources to train the same cumbersome model. In real federated learning settings, especially in cross-device FL settings where usually participating devices are small devices, this assumption might not be applicable. Thus, the participating clients might prefer to train a different model based on their resources or business needs.
As illustrated in Figure \ref{fig:3.8DLMH}, each client \( M_{\text{i}} \) has its own private data \(D_i=(X_i,Y_i )\) and a schema \( S_{\text{i}} \). As we assume that all clients would have only a few classes of data, thus each client model \( M_{\text{i}} \) would have only relevant target class nodes in the output layer. So, schema \( S_{\text{i}} \) contains the mapping of client classes w.r.t. global model classes. Each client \( M_{\text{i}} \) performs training on its personal private data \(D_i=(X_i,Y_i )\) and calculates the probability of only relevant classes. In DL-MH, to address the statistical heterogeneity challenge as well, each client is also asked to train a binary classifier C, which also has the same model architecture as the client model. Only the last layer of the client model is replaced with a 2-node output layer. Thus, it calculates a confidence matrix \textit{w} to measure the confidence value of each \(x\in X_{\text{dist}}\) for each client. Each client sends its logits \(\vartheta\), confidence matrix \textit{w}, and its schema \( S_{\text{i}} \)\((mask_{\text{dict}})\) to the server.
The server performs the mapping of client classes with global model classes and then applies masking on these mapped classes to make all clients’ outputs \(\vartheta\) of the same size.  Then, the server takes the weighted aggregation of clients’ knowledge by \( Y_g \leftarrow \Pi(\vartheta, w) \). Finally, the global model distils the knowledge from clients by training on \( X_{\text{dist}} \) using \( Y_{\text{g}} \)  as the target labels.

Figure \ref{fig:3.8DLMH} illustrates the overall working of DL-MH. To apply the network distillation objective in this scenario, this approach also extends the following distillation objective. 

\begin{equation}
\begin{aligned}
L(D_{\text{dist}}) &= \mathbb{E}_{x \sim x_{\text{dist}}} \left[l_1 \left( M_{\text{src}}(x), M_{\text{tgt}}(x) \right)\right] \\
&\quad + \mathbb{E}_{x, y \sim x_{\text{dist}} \times y_{\text{dist}}} \left[l_2 \left( M_{\text{tgt}}(x), y \right)\right]
\end{aligned}
\end{equation}

More specifically, this approach also provides the participating clients’ model M= \( M_{\text{i}} \) as a distillation source for the global model \( M_{\text{tgt}} \) where each client \( M_{\text{i}} \) is trained on non-IID data \(D_i=(X_i  ,Y_i )\). Furthermore, as explained earlier, the DL-MH approach also leverages the unlabeled public data as distillation data \(D_\text{dist}=X_\text{dist}\).  
\begin{algorithm}[H]
\caption{DL-MH (CLIENT TRAINING)}
\label{alg:dlmh_client}
\textbf{Description:} $M$ clients are indexed by $M_i$. $E$ is the number of local epochs for each client $M_i$. $C_i$ represents the binary classifier of client $i$. $E_{\text{embed}}$ is the number of epochs for training each binary classifier $C_i$. $M_{\text{tgt}}$ is the global model on the server, trained for $E_g$ epochs. $\eta$ is the learning rate. $D_i = (X_i, Y_i)$ is the local training data of client $M_i$. $D_{\text{dist}} = X_{\text{dist}}$ is the public unlabeled data for distillation. $F(\vartheta, \text{input})$ denotes the prediction function. $d$ represents the set of classes in $D_i$.

\begin{algorithmic}[1]
\For{each client $i \in M$}
    \State Create a dictionary \(mask_{\text{dict}}\) to save class mapping
    \State mask $\gets$ list of classes $d$
    \For{each class $j$ in $d$}
        \For{each sample $k$ in range of $Y_i$}
            \If{$Y_i[k] == d[j]$}
                \State $Y_i[k] \gets mask[j]$
            \EndIf
        \EndFor
        \State \(mask_{\text{dict}}[mask[j]]\) $\gets d[j]$
    \EndFor

    \State $B \gets$ split $D_i$ into batches of size $B$
    \For{each local epoch $e = 1$ to $E$}
        \For{each $b \in B$}
            \State $(y, x) \gets$ \Call{get\_train\_samples}{$b$}
            \State $\vartheta \gets \vartheta - \eta \nabla_\vartheta \{ \phi(y, F(\vartheta, x)) \}$
        \EndFor
    \EndFor

    \Comment{Train binary (embed) classifier $C_i$}
    \State $X_{\text{embed}} \gets$ concatenate$(X_i, X_{\text{dist}})$
    \State $Y_1 \gets$ zeros$(\text{len}(X_i))$
    \State $Y_2 \gets$ ones$(\text{len}(X_{\text{dist}}))$
    \State $Y_{\text{embed}} \gets$ concatenate$(Y_1, Y_2)$
    \State $D_{\text{embed}} \gets (X_{\text{embed}}, Y_{\text{embed}})$
    
    \State $B \gets$ split $D_{\text{embed}}$ into batches of size $B$
    \For{each local epoch $e = 1$ to $E_{\text{embed}}$}
        \For{each $b \in B$}
            \State $(y, x) \gets$ \Call{get\_train\_samples}{$b$}
            \State $w \gets w - \eta \nabla_w \{ \phi(y, F(w, x)) \}$
        \EndFor
    \EndFor
\EndFor
\end{algorithmic}
\end{algorithm}
Algorithm \ref{alg:dlmh_client} explains the working of clients’ training in DL-MH. Here, clients are expected to have solely different model architectures, including different target layers, which means clients could have only a limited number of target labels based on their available private training data. It means clients cannot directly start training on their private data; rather, they first need to map their target labels according to global target labels. More specifically, suppose client A has only two classes of data, say orange and banana, so it would have only two target labels: 0 for orange and 1 for banana. It implies that all training samples of client A would be mapped to 0 (orange) or 1 (banana). Similarly, client B might also have samples of two classes, say mango and orange, so here all training samples of client B would be mapped to 0 (mango) or 1 (orange).
Here it can be clearly observed that for both models, target labels are referring to their local classes rather than global classes (for the global model). Thus, for the global model, it would not be possible to take the simple aggregation based on clients’ local predictions. It is intuitive that the global model must perform the mapping and masking on the local predictions of clients before taking aggregation for the global model.
Therefore, to overcome this challenge, DL-MH assumes that each client already gets the global model target classes data and makes a mapping schema accordingly, as shown in Figure \ref{fig:3.8DLMH}. This assumption is absolutely simple and justifiable because, in real FL settings, it is intuitive that at the preprocessing stage, many things are negotiated/performed on the client device. Moreover, according to the literature review, in all simulations, clients are asked to train a model for a particular task and for given target classes. Like for CIFAR10, all clients are aware that they are going to train a deep learning model for image classification having 10 specific classes of data. To better understand the mapping and masking procedure, let's suppose client 1 has samples of classes 0, 3, 4, 7 whilst client N has samples of classes 4, 6, 9. Now, clients will map their class labels according to local numbering, which means client 1 will map its classes to 0, 1, 2, 3 and client N will map its class labels to 0, 1, 2. Now, clients will perform training on their private data using local mapped labels, as it can be seen in outputs produced by these clients in Figure \ref{fig:3.8DLMH}. Now, this locally mapped outputs are totally unknow to the server so clients send its mapping schema along with its output to the server. Server again apply the mapping on clients’ outputs using their own provided mapping schema. Furthermore, the server cannot perform direct aggregation on clients’ outputs as size of output vectors is different for each client and server needs a standard size of all clients’ outputs to take their average/aggregation. To overcome this challenge, server further applies the masking on these clients’ outputs where first it creates a new empty output vector of each client and put the client’s labels on relevant indexes and for remaining indexes, it put the value 0. For instance, client 1 has samples of classes 0,3,4,7 so its output values are put on relevant indexes 0, 3, 4 and 7 and all other indexes have value 0. 
 In Algorithm \ref{alg:dlmh_client}, lines 1-11, all clients create a mapping schema ‘\( mask_{\text{dict}} \)’ and update their target labels accordingly. In lines 13-18, all clients train their local models on their own private data \( D_{\text{i}} \)=(\( X_{\text{i}} \),\( Y_{\text{i}} \) ). All clients train their model for E number of epochs and return the logits \(\theta\). Similar to first proposed approach, DL-SH, this approach, DL-MH, also adaptively aggregates the client models and then these adaptively calculated outputs are used to train the global model. More specifically, rather than performing simple aggregation on client models’ output, the proposed approach introduces the concept of confidence matrix w which basically informs how much a client is confident about its predictions against provided classes. 
To calculate the confidence matrix \( W_{\text{i}} \) (x) with unlabeled data, each client is asked to train a binary classifier \( C_{\text{i}} \) (x)=[0,1] with sigmoid outputs. The objective of this classifier is to distinguish the client’s local private data  \( X_i \in D_i \) and \( X_{\text{dist}} \in D_{\text{dist}} \)
. As shown in Figure 3.6, each client’s binary classifier is created by modifying the client's local model. More specifically, only the last layer of the client's local model is replaced by 2 nodes dense layer. The classifier behaves much similar to the discriminators trained in a Generative Adversarial Network \cite{Goodfellow2014}. It is assumed that each client i trained its binary classifier \( C_{\text{i}} \) optimally, thus the prediction of \( C_{\text{i}} \) (x) can be expressed as follows:
Here \( p_{\text{i}} \)(x) is the probability of sample (x) on  \( X_{\text{i}} \) and  \( p_{\text{dist}} \) (x) is the probability of sample (x) on \( X_{\text{dist}} \).
By fixing the value of x and considering \( p_{\text{dist}} \) (x) as a positive constant, \(C_i^* \) (x) monotonically increases with  \( p_{\text{i}} \) (x) within  \( p_{\text{i}} (x)\in[0,1] \) . It implies that \(C_i^*\) would compute higher values when sample  x is more likely to be present in \( X_{\text{i}} \). Thus client \( M_{\text{i}} \) would be more confident about its prediction \( M_{\text{i}} \) (x).
More technically, after completing the training of local models on their private data, each client trains its own binary classifier \( C_{\text{i}} \) on the client’s private data \( X_{\text{i}} \) and public data \( X_{\text{dist}} \).  To train the binary classifier, in lines 11-15, each client first combines the client's private training samples along with the public unlabeled training samples by 
\( X_{\text{embed}}\leftarrow\text{concatenate}(X_{\text{i}},X_{\text{dist}}) \)
For target variables, training data is assigned as label 0 and public unlabeled data is assigned as 1. Finally, these target labels are combined as well:\\
\\
\( Y_1 \leftarrow \text{zeros}(\text{len}[X_i]) \)\\
\\
\( Y_2 \leftarrow \text{ones}(\text{len}[X_\text{dist}]) \)\\
\\
\( Y_\text{embed}\leftarrow \text{concatenate}(Y_1  ,Y_2) \)
\\
\\
So, now, there is a new dataset for binary classifier training i.e.
\( D_\text{embed} \leftarrow (Y_\text{embed}  ,Y_\text{embed}) \)
In lines 20-30, the binary classifier \( C_{\text{i}} \) performs the training on \( D_\text{embed} \leftarrow (Y_\text{embed}  ,Y_\text{embed}) \) for \(E_{\text{embed}} \) number of epochs. After training, the binary classifier \(C_{\text{i}} \) returns the confidence matrix w.\\
After completing the client’s training, the server starts its distillation and training process as indicated in Algorithm \ref{alg:dlmh_server}. Here, as stated earlier, the server cannot directly perform aggregation and distillation on clients’ provided knowledge (logits, confidence matrix and mapping schema) due to different class labels mapping. Therefore, the server first performs the mapping on each client’s logits to make it in accordance with global model class labels. In lines 3-9, for each client, mapping and masking are performed where first all clients’ logits are mapped to global model target labels according to the client’s mapping schema. Then these mapped logits are masked in ‘\(w_{\text{pred}}\)’ (a list of zeros having size = client’s logit * target labels of the global model) as illustrated in Figure \ref{fig:3.8DLMH}.

\begin{algorithm}[H]
\caption{DL-MH (SERVER TRAINING)}
\label{alg:dlmh_server}
\textbf{Description:} $M$ clients are indexed by $M_i$. $E$ is the number of local epochs for each client $M_i$. $C_i$ is the binary classifier of client $i$. $E_{\text{embed}}$ is the number of epochs for each binary classifier $C_i$. $M_{\text{tgt}}$ represents the global model on the server, trained for $E_g$ epochs. $\eta$ is the learning rate. $D_i = (X_i, Y_i)$ is the local dataset of client $M_i$. $D_{\text{dist}} = X_{\text{dist}}$ is the public unlabeled dataset used for distillation. $F(\vartheta, \text{input})$ is the prediction function. $\theta$ is the model prediction. $d$ is the set of class labels in $D_i$.

\begin{algorithmic}[1]
\For{each client $i \in M$}
    \State Create a list of zeros \(w_{\text{pred}}\) of size $|\theta_i| \times \texttt{total\_classes}$
    \For{each sample $k$ in range of $\theta_i$}
        \State \texttt{index} $\gets 0$
        \For{each class $j$ in $d$}
            \State ${w_{\text{pred}}}[k][j] \gets \theta[k][\texttt{index}]$
            \State \texttt{index} $\gets$ \texttt{index} $+ 1$
        \EndFor
    \EndFor
    \State $\vartheta_i \gets$ ${w_{\text{pred}}}$
    \State $w_i \gets \frac{\exp(w_i/T)}{\sum_j \exp(w_j/T)}$
\EndFor

\State $\vartheta \gets \sum\limits_{i=1}^{M} \vartheta_i$
\State $w \gets \sum\limits_{i=1}^{M} w_i$
\State $Y_g \gets \Pi(\vartheta, w)$
\State $D_g \gets (X_{\text{dist}}, Y_g)$

\State $B \gets$ split $D_g$ into batches of size $B$
\For{each global epoch $e = 1$ to $E_g$}
    \For{each $b \in B$}
        \State $(y, x) \gets$ \Call{get\_train\_samples}{$b$}
        \State $w \gets w - \eta \nabla_w \{ \phi(y, F(w, x)) \}$
    \EndFor
\EndFor
\end{algorithmic}
\end{algorithm}

In line 11, the server applies the softmax(w).   When client model \(M_i\) is trained on non-IID data and provides a variety of responses to \(x \in X_{\text{dist}}\), then confidence \(w_i\) (x) should be higher for that client who has observed more similar samples of the same classes in their private training data. Thus, this client would be more confident to inform the global model about the output of sample x. Confidence matrix \(w_i\) (x) is calculated by applying the softmax activation function on \(C_i\) (x) as follows:
\begin{equation}
W_i(x) = \frac{\exp\left( C_i(x / T) \right)}{\sum_j \exp\left( C_j(x / T) \right)}
\label{eq:3.6}
\end{equation}

Here the purpose of applying the softmax is to ensure that \( \sum_i w_i(x) = 1 \). Moreover, T is the temperature parameter to control the smoothness of the output. As mentioned earlier, temperature T has a very significant impact while transferring knowledge from one end to another. 
After applying softmax, the server performs the aggregation on the client’s logit \(\theta\)  and confidence matrix w in lines 13 and 14 respectively. Then server multiplies the clients’ logits \(\theta\)  with their confidence matrix w  to compute target labels \(Y_g\). Then, the global model \(M_{\text{tgt}}\)  used this \(Y_g\) along with public unlabeled data \(X_{\text{dist}}\) as follows:
\(D_g\leftarrow (X_{\text{dist}}  ,Y_g )\)
Finally, the server trains its global model \(M_{\text{tgt}}\)  on \(D_g\) for E number of epochs.
\subsection{Client Incentive-Based Decentralized Learning Approach To Leverage Heterogeneous Clients With Non-Iid Data}	
\begin{figure*}[t]
    \centering
    \includegraphics[width=\linewidth]{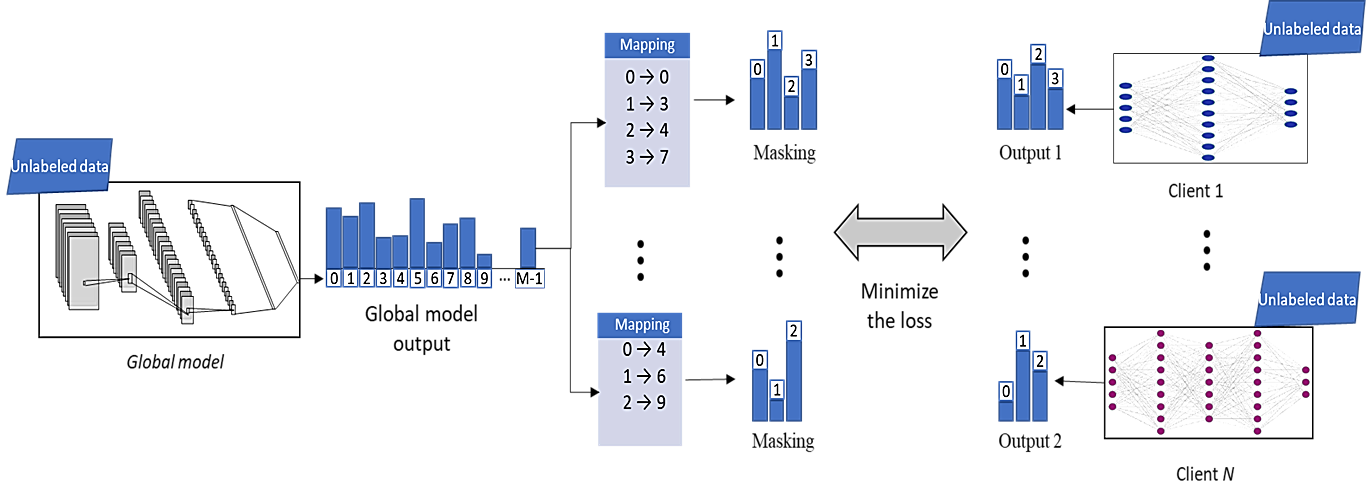}
    \caption{Overview of I-DL-MH}
    \label{fig:3.11I-DL-MH}
\end{figure*}

This paper further extends the second proposed approach named DL-MH by enabling the clients to also get incentives in the form of updated knowledge from the server. The extended approach is titled Incentive-based Decentralized Learning with Model Heterogeneity (I-DL-MH). The proposed approach, I-DL-MH, very efficiently addresses the client incentives challenge where clients almost have negligible additional computational or computational overhead while distilling the knowledge from the server. Figure \ref{fig:3.11I-DL-MH} illustrates the overall working of I-DL-MH.
The primary objective of federated learning is to train an efficient global model by leveraging the confidential data of participating clients. The proposed approaches DL-SH and DL-MH have already effectively achieved this objective under various challenging conditions. In federated learning settings, typically, it is assumed that clients would be voluntarily convinced to participate in federated learning training however, in practical scenarios, it could be very difficult to convince the clients to allow someone to consume their computational and communicational resources along with valuable data with no incentives. In pioneer work \cite{BrendanMcMahan2017,Konecny2016}of federated learning by Google, they use the complete model sharing approach where in each round, participating clients can also get the updated trained model from the server (as incentives) so they can also use that updated model on their devices. Moreover, Google has access to (android) operating systems (OS) of millions of devices, so they can return the incentives in the form of OS updates. However, for small third-party developers/organizations, it might not be possible to give incentives to participating clients in the same manner. Particularly, under fully model-heterogeneous settings, this could be more difficult where all clients may have entirely different model architectures, including different targets. Thus, the client cannot distil the knowledge from the global model straightforwardly.
Here, this paper proposes a client incentive-based approach named Incentives Based Decentralized Learning with Model Heterogeneity (I-DL-MH) by further extending the proposed approach DL-MH to address this issue. As explained earlier, DL-MH allows the fully heterogeneous models to effectively collaborate to train a global model at the server’s end. During the training, clients also provide their mapping schema \(S_i(mask_{\text{dict}})\) to the server, which leverages this  schema to mask the client’s local classes on the global model’s target classes. Then the server performs the global model distillation on the updated weighted logits of clients using public unlabeled data \(X_\text{dist}\). 
In the I-DL-MH approach, after performing the global model training, the server also shares its logits with all participating clients who are interested to get the incentive in the form of updated knowledge (global model learning).  Here, the server needs to perform mapping and masking on their logits according to each client data distribution to make it in accordance with each client model’s target labels. To reduce the computational overhead on the client device, this step is performed on the server where it is assumed that the server typically has reasonably large computational and communicational resources. These individually mapped and masked logit matrices of the server are shared with each client accordingly. To further reduce the communicational overhead on the client device, clients can leverage the same public unlabeled data \(X_\text{dist}\) to perform the distillation on the server logits.

Thus, I-DL-MH can enable the participating heterogeneous clients to get incentives in the form of updated knowledge from the global model. In this approach, the student model gets almost 225\% performance gain with only one round to distil from the global model under the most complex data distribution (non-IID) in the CIFAR10 dataset.

As explained earlier, DL-MH allows the fully heterogeneous models to effectively collaborate to train a global model at the server’s end. During the training, clients also share their mapping schema \(S_i(mask_{\text{dict}}\) ) with the server and server leverages this mapping schema to mask the client’s local classes on the global model’s target classes. Then server performs the global model distillation on the updated weighted logits of clients using public unlabeled data \(X_{\text{dist}}\).
In the I-DL-MH approach, after performing the global model training, the server also shares its logits with all participating clients who are interested to get the incentive in the form of updated knowledge (global model learning).  Here, the server need to perform mapping and masking on their logits according to each client data distribution to make it in accordance with each client model’s target labels.

\begin{algorithm}[H]
\caption{I-DL-MH (SERVER TRAINING)}
\label{alg:idl_mh_server}
\textbf{Description:} $M$ clients are indexed by $M_i$. $E$ is the number of local epochs. $M_{\text{tgt}}$ is the global model trained for $E_g$ epochs. $\eta$ is the learning rate. $D_g = (X_{\text{dist}}, Y_g)$ is the public distillation dataset formed by aggregating client logits: $Y_g \gets \Pi(\vartheta, w)$. $F(\vartheta, \text{input})$ is the prediction function. $\theta$ is the model prediction.

\begin{algorithmic}[1]
\State $D_g \gets (X_{\text{dist}}, Y_g)$
\State $B \gets$ split $D_g$ into batches of size $B$
\For{each global epoch $e = 1$ to $E_g$}
    \For{each $b \in B$}
        \State $(y, x) \gets$ \Call{get\_train\_samples}{$b$}
        \State $\vartheta \gets \vartheta - \eta \nabla_\vartheta \{ \phi(y, F(\vartheta, x)) \}$
    \EndFor
\EndFor

\For{each client $i \in M$}
    \For{each sample $k$ in range of $\theta$}
        \State $max \gets \max(\theta[k])$
        \State $min \gets max - \theta[k][0]$
        \State $min\_idx \gets d[0]$
        \For{each class $j$ in $d$}
            \If{$(max - \theta[k][j]) < min$}
                \State $min \gets max - \theta[k][j]$
                \State $min\_idx \gets j$
            \EndIf
        \EndFor
        \State $\theta[k] \gets \mathbf{0}$
        \State $\theta[k][min\_idx] \gets max$
    \EndFor

    \State $counter \gets 0$
    \State $mask \gets$ \Call{list}{$|d|$}
    \State $unique\_labels \gets$ \Call{unique}{$d$}

    \For{each $j$ in $|d|$}
        \If{$d[j] \in unique\_labels$}
            \For{each sample $k$ in range of $\theta$}
                \If{$\theta[k] == unique\_labels[counter]$}
                    \State $\theta[k] \gets mask[j]$
                \EndIf
            \EndFor
            \State $counter \gets counter + 1$
        \EndIf
    \EndFor
\EndFor
\end{algorithmic}
\end{algorithm}

Algorithm \ref{alg:idl_mh_server} explains the working of server training/processing in I-DL-MH. In line 1, it is assumed that the server has already received the confidence matrices and client logits from clients and now the server has created a dataset \(D_g\leftarrow (X_{\text{dist}}  ,Y_g )\). In line 2-8, global model performs adaptive distillation using the \(D_g\). After finishing the global model training, server needs to transform its global model’s logits in accordance with each client’s data distribution by applying some mapping and masking techniques. As the global model logits contain values of all target labels while clients would most probably contain only a few of these target labels. Thus, to transform these complete logits in accordance with the client’s logits, the server performs some transformations in lines 10-22. Here it is possible that a sample \(x\in X_{\text{dist}}\) might have max logit value against a class which is not available at the client \(M_i\) so I-DL-MH finds the nearest available client class and assigns the maximum logit value to that class.  It helps the client to map the sample x according to their best available class. Empirically, it was observed that this approach gives very promising results for every client having absolutely distinct data distribution. 
In lines 24-36, the server applies the mapping and masking on updated global model logits according to each client’s mapping schema ‘\(mask_{\text{dict}}\)’ or d. To reduce the computational overhead on the client device, this step is performed on the server where it is assumed that the server typically has reasonably large computational and communicational resources. These individually mapped and masked logit matrices of the server are shared with each client accordingly. To further reduce the communicational overhead on the client device, clients can leverage the same public unlabeled data \(X_{\text{dist}}\) to perform the distillation on the server logits.
\begin{algorithm}[H]
\caption{I-DL-MH (CLIENT-TRAINING)}
\label{alg:client_training}
\textbf{Description:} $M$ clients are indexed by $M_i$. $E$ is the number of local epochs. $C_i$ is the binary classifier of client $i$. $E_{\text{embed}}$ is the number of epochs for the binary classifier. $M_{\text{tgt}}$ is the global model on the server, trained for $E_g$ epochs. $\eta$ is the learning rate. $D_i' = (X_{\text{dist}}, Y_g')$ is the distillation dataset for client $M_i$. $F(\vartheta, \text{input})$ is the prediction function.

\begin{algorithmic}[1]
\For{each client $i \in M$}
    \State $D_i' \gets (X_{\text{dist}}, \theta_i)$
    \State $B \gets$ split $D_i'$ into batches of size $B$
    \For{each local epoch $e = 1$ to $E$}
        \For{each $b \in B$}
            \State $(y, x) \gets$ \Call{get\_train\_samples}{$b$}
            \State $w \gets w - \eta \nabla_w \{ \phi(y, F(w, x)) \}$
        \EndFor
    \EndFor
\EndFor
\end{algorithmic}
\end{algorithm}

Algorithm \ref{alg:client_training} explains the client training after receiving the global model logits from the server. In this scenario, as there is no concept of a binary classifier, so each client directly receives its transformed and modified global model logits \(\theta_{\text{i}}\) and use the already downloaded public data \(X_{\text{dist}}\) along with it to make a new dataset \( D_i' = (X_{\text{dist}}, \theta_i) \)
 . Then each client starts distillation and further training on \(D_i'\) for some number of epochs.
I-DL-MH shares the logits(\(\theta\))  of the global model \(M_{\text{tgt}}\), after some processing, with the clients so that clients can also update their own model by distilling from the updated knowledge. However, it was observed that weighted aggregation \(Y_{\text{agg}}\) can also be leveraged to distil the knowledge from the global model to clients. Initial experiments demonstrated that clients almost get the same performance gain in both scenarios. 
I-DL-MH doesn’t put extra communication overhead for clients to distil the updated knowledge from the global model. More specifically, now, clients are not required to download or receive more unlabeled public data to perform the distillation because clients already have downloaded/received \(D_{\text{dist}}=(X_{\text{dist}} )\) while initial training (DL-MH). Thus, the same unlabeled public dataset can be leveraged to distil the knowledge from the global model. Hence, in such a way, clients can get their incentives in the form of updated knowledge from the global model with almost negligible additional communication overhead.
Contrary to this, in the DL-SH scenario, this incentive approach is comparatively easy to deploy as all clients have the same target labels thus server doesn’t need to do additional processing in the form of mapping and masking to make data relevant and compatible for each client. In the DL-SH scenario, clients only need to get the weighted aggregation \(Y_{\text{agg}}\) or the logits (\(\theta)\)  of the global model from the server and then they can distil the updated knowledge from the global model.

\section{Experimental setup}
\label{sec:Experimental setup}
\subsection{Datasets}
To evaluate the performance of the proposed approaches, this work considers the following datasets as benchmark datasets. According to the literature, these are the most commonly used datasets in the FL setting. Among these datasets, CIFAR100 and CINIC10 are more challenging and complex datasets to evaluate the performance of deep learning models. The proposed approaches/solution in this work are based on \cite{Ma2020} which requires distinct training data referred to as distillation data  \( (X_{\text{dist}}, Y_{\text{dist}}) \). Distillation data is usually extracted from primary training data \( (X_{\text{train}}, Y_{\text{train}}) \). 
\begin{table}[H]
\caption{Data Split of Benchmark Datasets}
\label{table:data_split}
\centering
\begin{tabular}{|l|c|c|c|}
\hline
\textbf{Dataset} & \textbf{Training Data} & \textbf{Distillation Data} & \textbf{Client Data} \\
\hline
MNIST           & 48000 & 12000 & 10000 \\
Fashion MNIST   & 48000 & 12000 & 10000 \\
CIFAR10         & 40000 & 10000 & 10000 \\
CIFAR100        & 40000 & 10000 & 10000 \\
CINIC10         & 90000 & 90000 & 90000 \\
\hline
\end{tabular}
\end{table}
Since in real-world scenarios, it is intuitive that unlabeled data is more easily available as compared to labelled/annotated data, therefore, this paper splits each training data in an 80\%-20\% ratio for distillation data. More specifically, it randomly chose 80 \% of samples from training data as distillation data and the remaining samples are used as training data (client data pool) for training the local models on devices. For CINIC10, it naturally includes a training dataset and validation dataset comprising 90000 samples each thus this paper leverages the validation dataset as distillation data \( X_{\text{dist}}\). Table \ref{table:data_split} shows the data split of benchmark datasets in client, training, and distillation data. Each device has its predefined class probability \( p_{\text{i}}\). Therefore, each device creates its own training data \( D_{\text{i}}\) =(\(X_{\text{i}} ,Y_{\text{i}}\)) from the client data pool using its class probability \( p_{\text{i}}\). To ensure that each device has sufficient data samples to train deep networks, each device was supposed to have its local dataset of half the size of the client data pool. More specifically, for MNIST and FMNIST, devices were supposed to create their local dataset of size 6000 allowing duplicates. For CIFAR10 and CIFAR100, devices created their dataset of size 5000 allowing duplicates. However, for CINIC10, each device created its dataset of size 20000 allowing duplicates.
\subsection{Data Distributions}
This paper considers four different data distribution settings of varying difficulty levels to evaluate the performance of the proposed approaches. Basically, there are two primary divisions as IID data distribution and non-IID data distribution. In IID settings, it is assumed that the data distributed is independent and identical among all participating devices. As discussed earlier, it is a very strong and unrealistic assumption in a real-world scenario. It might be possible only in a centralized and controlled environment where data is distributed manually. Whilst in non-IID, it is assumed that data could be unbalanced and non-identical. To determine the robustness of proposed approaches, this paper further distributes the non-IID into 3 different types of non-IID data distributions named non-IID 1(NIID-1), non-IID 2(NIID-2) , and non-IID 3(NIID-3) . In these settings, each class has different device-wise class probabilities. To explain these settings, this work assumes 5 devices and a dataset of 10 classes, p represents the class probability for each device i.
\begin{itemize}
\item \textbf{IID}

In IID data distribution, all devices have equal class probabilities i.e.,

\( p_i=[0.1,0.1,…,0.1]\) 
means all devices have samples of all classes.

\item \textbf{non-IID 1 (NIID-1)}

In NIID-1, each device holds samples of only two consecutive classes. It is the most challenging setting for devices. In such a scenario,
\begin{align*}
p_{5n+1} &= [0.5,\, 0.5,\, 0,\, 0,\, \ldots,\, 0] \\
p_{5n+2} &= [0,\, 0,\, 0.5,\, 0.5,\, 0,\, \ldots,\, 0] \\
p_{5n+3} &= [0,\, 0,\, 0,\, 0,\, 0.5,\, 0.5,\, \ldots,\, 0]
\end{align*}
and so on.
 
\item \textbf{non-IID 2 (NIID-2)}
In NIID-2, all devices commonly share samples of the first 5 consecutive classes and each device also holds samples of one unique class. In this scenario, 
\begin{align*}
p_{5n+1} &= [1/6,1/6,1/6,1/6,1/6,1/6,0,0,0,0] \\
p_{5n+2} &= [1/6,1/6,1/6,1/6,1/6,0,1/6,0,0,0] \\
p_{5n+3} &= [1/6,1/6,1/6,1/6,1/6,0,0,1/6,0,0]
\end{align*}
and so on.

\item \textbf{non-IID 3 (NIID-3)}
In NIID-3, each device shares samples of only 4 classes however in such a way that samples of each class are distributed between two devices only. In such a scenario,
\begin{align*}
p_{5n+1} &= [0.25,\, 0.25,\, 0.25,\, 0.25,\, 0,\, \ldots,\, 0] \\
p_{5n+2} &= [0.25,\, 0,\, 0,\, 0,\, 0.25,\, 0.25,\, 0.25,\, \ldots,\, 0] \\
p_{5n+3} &= [0,\, 0.25,\, 0,\, 0,\, 0.25,\, 0,\, 0,\, 0.25,\, 0.25, 0]
\end{align*}
and so on.
\end{itemize}
\subsection{Models}	
To evaluate the performance of the proposed approaches, this paper uses the mostly used and state of art deep learning models. Most of the work\cite{Ma2020,He2020b,Liang2018,Vongkulbhisal2019} in FL have used ResNet and DenseNet as deep learning models to evaluate the performance of their proposed approaches. ResNet18 \cite{He2016} is a very popular and state of art deep learning model for classification tasks which has outperformed many traditional machine learning models in computer vision tasks. Resnet18 has 18 layers. DenseNet \cite{Huang2017} is another state of art deep learning model which has fewer parameters than Resnet18. To evaluate the robustness of proposed approaches, this paper performs extensive experiments on various deep learning models of different architectures (different depths/complexities). Though, this paper has leveraged all deep learning models to evaluate the performance of proposed approaches however, for better understanding and comparison, this paper  labeled the models based on their comparative complexities like this research work considers ResNet18 as a complex/deep model as it has many hidden layers and DenseNet as a shallow model as comparatively, it has less hidden layers or have small architecture than ResNet18. To further check the robustness of proposed approaches on small devices having very low computational resources, this paper also considers another very small model named ResNet8. It is essentially a modification of ResNet18 and has only 8 layers. 

\subsection{Baseline and Evaluation Metrics}
In literature, mostly research works \cite{Anil2018,He2020b,Ma2020} have considered the standard FL (FedAvg) as a baseline. There are many other relevant works in FL however these were not applicable in the supposed FL settings of this paper and thus were not used as the baseline. For instance, FedProx \cite{Li2018} performs worst for deep networks even worse than FedAvg as demonstrated by \cite{Wang2020}. Similarly, FedMA \cite{Wang2020} does not support batch normalization layers thus not possible to employ in supposed FL settings where ResNet is used as a deep learning model \cite{He2020b}. The closest works related to this work are FedGKT \cite{He2020b} and UHC \cite{LUCASCASTELLANO2021,Vongkulbhisal2019}. Though in UHC, authors have not implemented their approach for FL settings, and they have just implemented their approach at a small scale to show that in their approach students can distil the knowledge from the teacher model regardless of varying target labels. However, this paper believes that their work could be further extended with some efforts to employ in FL settings. Thus, UHC was first extended according to FL settings before making a comparison with it. Another work ODIN \cite{Liang2018} has shown that it can effectively address the out-of-distribution (non-IID) problem thus this paper also uses this approach as a benchmark. This paper also implements and uses the true labelled \((global_\text{true})\) approach to compute and benchmark the upper bound performance of the global model in given FL settings. Thus, this work considers many benchmarks to evaluate the effectiveness of proposed approaches.
To evaluate the performance of proposed approaches, this paper considers the various evaluation metrics according to the problem context. Most of the FL literature on image classification tasks including benchmark approaches \cite{BrendanMcMahan2017,Vongkulbhisal2019,Wang2020,He2020b,Ma2020} have employed the model test accuracy as the evaluation metric to evaluate the performance of their proposed approaches. Thus, this work also considers the model test accuracy as a primary evaluation metric. However, as one objective of this work is also concerned with communication overhead hence, to evaluate the proposed solution , this work considers the communication cost as the evaluation metric as well.
To calculate the communication cost, this work employs the method used in \cite{He2020b} where the authors calculate the communication cost for one client using the following equation.

\begin{equation}
\begin{split}
\text{Comm\_cost} = \left( \text{hidden\_features} + \text{logits\_server}+ \text{logits\_client}\right) \\
\times (\text{dataset\_size}) \times (\text{comm\_rounds}) \times M
\end{split}
\label{eq:3.7}
\end{equation}
This work computes the communication cost of proposed approaches by 

\begin{equation}
\begin{split}
\text{Comm\_cost}_{\text{FedAvg}} = \left(\text{model\_params}_{\text{server}} + \text{model\_params}_{\text{client}}\right) \\
\times (\text{number\_epochs})  \times M
\end{split}
\label{eq:3.8}
\end{equation}
\begin{equation}
\text{Comm\_cost}_{\text{DL-SH}} = \left( X_{\text{dist}} \times \vartheta + w \right) \times M
\label{eq:3.9}
\end{equation}

\begin{equation}
\text{Comm\_cost}_{\text{DL-MH}} = \left( X_{\text{dist}} \times \vartheta + w + \text{mask\_dict} \right) \times M
\label{eq:3.10}
\end{equation}

Here
M denotes the number of clients.
\(X_\text{dist}\) denotes the public unlabeled data
\(\theta\) denotes the logit size
w denotes the confidence matrix size

Eq. \ref{eq:3.8} calculates the communication cost of standard FedAvg algorithm where there are no \(hidden\_features\) and it is independent of dataset size because here M clients exchange their parameters \(model\_params_{\text{client}} \) with the server \(model\_prarms_{\text{server}}\) for \(number\_epochs\) communication rounds. Eq. \ref{eq:3.9} calculates the communication cost of first proposed approach (DL-SH) where M clients share the confidence matrices w along with logits \(\theta\) against each sample from public unlabeled data \(X_{\text{dist}}\). Similarly, Eq. \ref{eq:3.10} calculates the communication cost of second proposed approach (DL-MH) where only relevant logits \(\theta\) of clients are shared along with the mapping schema \(mask_{\text{dict}}\).
\section{Result and Discussion:}
This  section present the performance of proposed approaches on different data distributions with different datasets and using different model architectures. Very large-scale experiments were performed to evaluate the effectiveness of the proposed approaches. Keeping in view the limitations of the length, only very few experimental results are presented here for reference.  

\subsection{DL-SH}	
\begin{figure}[H]
    \centering
    \includegraphics[width=1\linewidth]{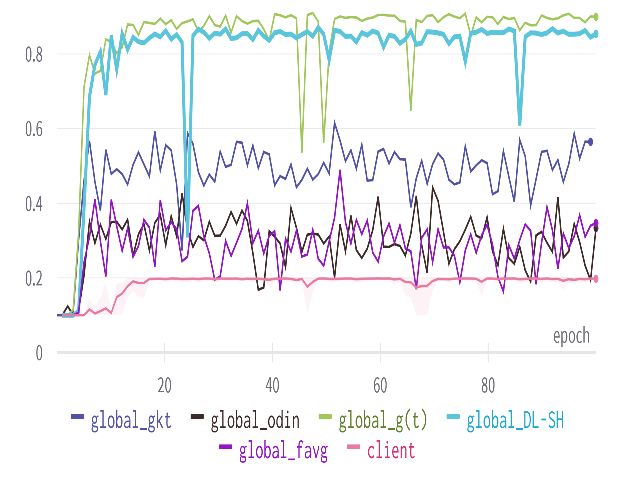}
    \caption{Test accuracy on FMNIST NIID-1 data distribution using model$_{global}$ = RN18 \& model$_{locals}$ = RN18}
    \label{fig:4.2FMNIST IID}
\end{figure}
\begin{figure}[!ht]
    \centering
    \includegraphics[width=0.8\linewidth]{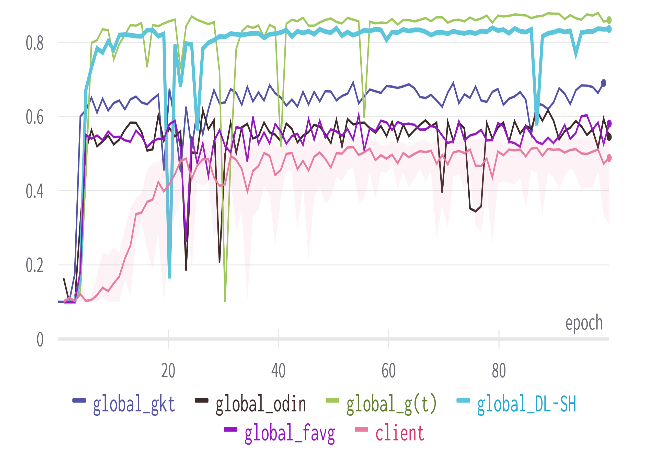} \\
    \vspace{0.5em}
    \includegraphics[width=0.8\linewidth]{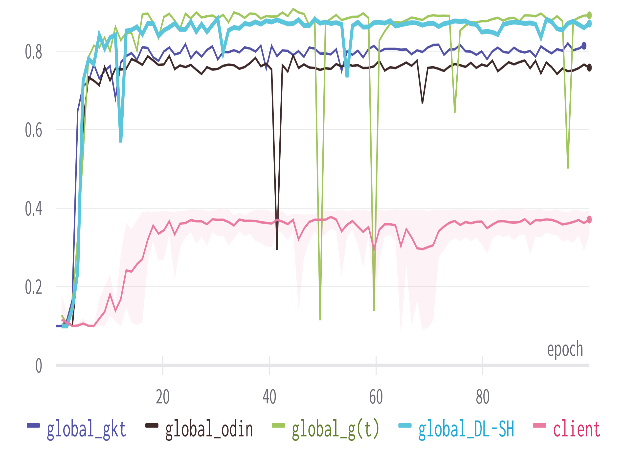}
    \caption{Test accuracy on FMNIST NIID-2(first) and NIID-3(second) data distribution using model$_{global}$ = RN18 \& model$_{locals}$ = RN18}
    \label{fig:4.3FMNIST non-IID}
\end{figure}

 In Figure \ref{fig:4.2FMNIST IID}  it can be observed that clients have around 0.2 test accuracy in the NIID-1 scenario, which is intuitive because all clients have training samples of only 2 classes (out of 10). Similarly, in Figure \ref{fig:4.3FMNIST non-IID} NIID-2 scenario clients (clients-avg) perform better than the NIID-1 and NIID-3 cases. This behaviour is intuitive because in the NIID-2 case, all clients have samples of 5 common classes, and only one class is unique for each client. 

However interestingly, in the NIID-3 cases, all approaches (global models) perform better than NIID-1 and NIID-2 scenarios. One potential reason could be that in NIID-3 data distribution, each class’s samples are distributed among at least 2 clients means at least two clients try to learn the features of each class. This argument is also supported by the fact the in IID scenario, all clients try to learn the features of each class so all SOTA performs best in the IID case as compared to other data distributions. 

\begin{figure}[!ht]
    \centering
    \includegraphics[width=0.8\linewidth]{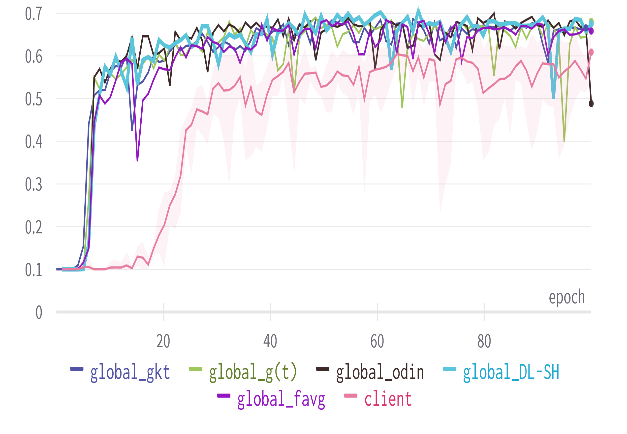} \\
    \vspace{0.5em}
    \includegraphics[width=0.8\linewidth]{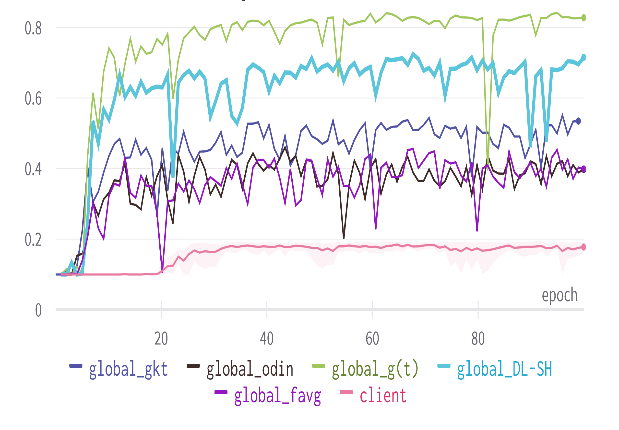}
    \caption{Test accuracy on CIFAR10 IID (first) and NIID-1 (second) data distribution using model$_{global}$ = RN18 \& model$_{locals}$ = RN18}
    \label{fig:4.4CIFAR10 IID/noniid}
\end{figure}

\begin{figure}[!ht]
    \centering
    \includegraphics[width=0.8\linewidth]{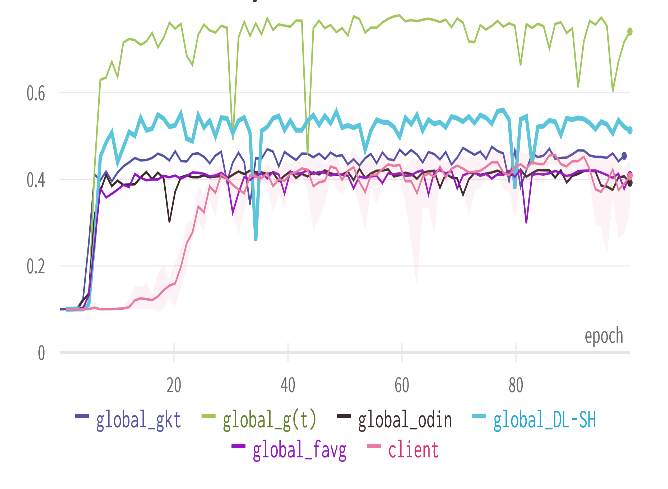} \\
    \vspace{0.5em}
    \includegraphics[width=0.8\linewidth]{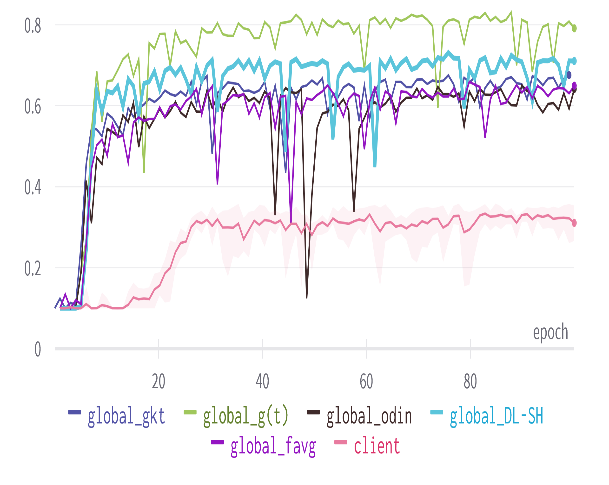}
    \caption{Test accuracy on CIFAR10 NIID-2(first) and NIID-3(second) data distribution using model$_{global}$ = RN18 \& model$_{locals}$ = RN18}
    \label{fig:4.5CIFAR10 non-IID}
\end{figure}

Nevertheless, it can be observed clearly, that in all these scenarios, the proposed approach again performs much better than SOTA approaches.
Almost similar behavior can be seen in Figure \ref{fig:4.4CIFAR10 IID/noniid} and Figure \ref{fig:4.5CIFAR10 non-IID} against the CIFAR10 dataset. 
\subsubsection{Clients as a shallow model (DenseNet)}
\begin{figure}[!ht]
    \centering
    \includegraphics[width=0.8\linewidth]{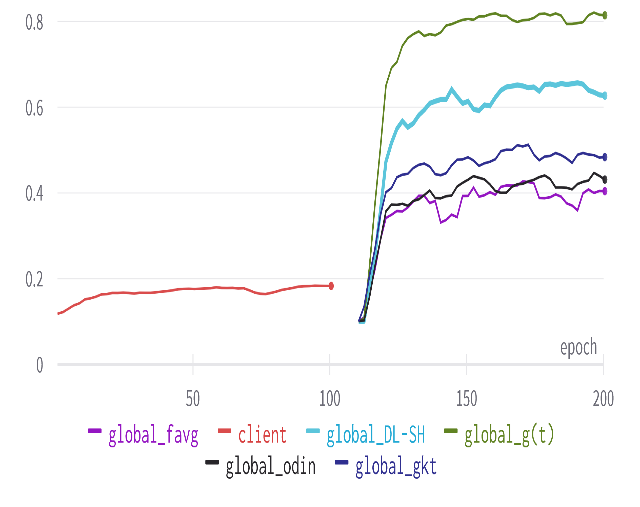} \\
    \vspace{0.5em}
    \includegraphics[width=0.8\linewidth]{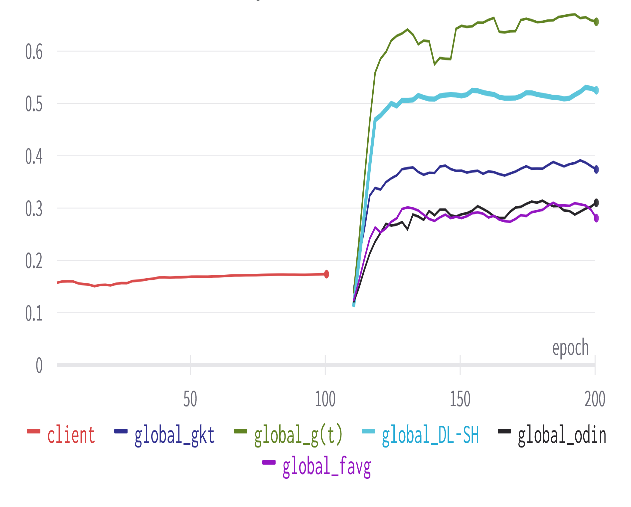}
    \caption{Test accuracy on CIFAR10 NIID-1 (first) and CINIC10 NIID-1 (second) data distribution using model$_{global}$ = RN18 \& model$_{locals}$ = DN}
    \label{fig:4.8CIFAR10 non-IID}
\end{figure}

Figure \ref{fig:4.8CIFAR10 non-IID} illustrates the performance of global models when all clients are using a shallow model architecture (DenseNet).  Here, it can be observed that the performance of all SOTA approaches reasonably degrades when all clients train the shallow model as compared to Figure \ref{fig:4.4CIFAR10 IID/noniid} where all clients are using deep model architectures for the CIFAR10 dataset with NIID-1 data distribution. Moreover, it can also be observed that the performance of clients along with SOTA approaches decreased in the case of the CINIC10 dataset as compared to the CIFAR10 dataset. It is intuitive because CINIC10 is a more complex and larger dataset as compared to the CIFAR10 dataset.

\subsubsection{Clients as shallow/deep (hybrid) model (DenseNet/ ResNet18}
From Figure \ref{fig:4.4CIFAR10 IID/noniid}, it can be observed that, in most of the scenarios, the hybrid client model architecture approach provides better results than all-client-models as a shallow model but slightly lower than all-client-models as a deep model. 

\subsection{DL-MH}
\begin{figure}[!ht]
    \centering
    \includegraphics[width=0.8\linewidth]{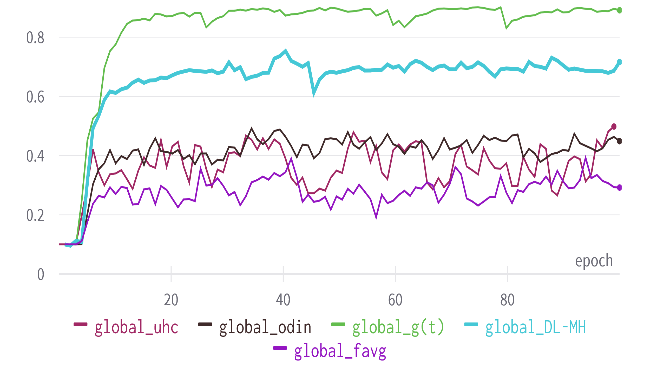} \\
    \vspace{0.5em}
    \includegraphics[width=0.8\linewidth]{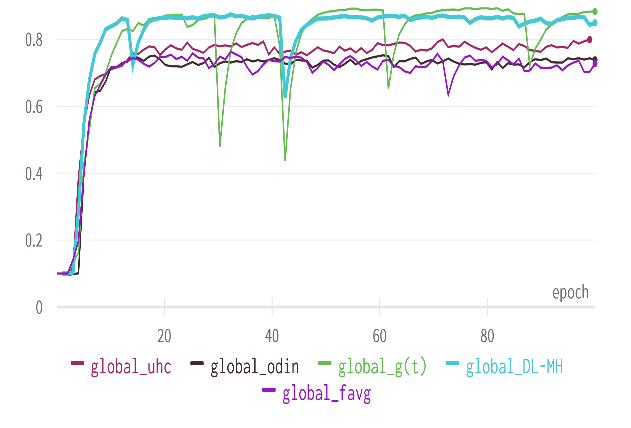}
    \caption{Test accuracy on FMNIST- NIID-1 (first) and NIID-3 (second) data distribution using model$_{global}$ = RN18 \& model$_{locals}$ = RN18}
    \label{fig:4.12 FMNIST non-IID}
\end{figure}

In addition to other metrics, this section also shows the communication efficiency of DL-MH as compared to standard FL, simple distillation approach. It also compares the communication efficiency of DL-MH with the first proposed approach DL-SH to show that the current approach is much more cost-effective as compared to DL-SH. 
In Figure \ref{fig:4.12 FMNIST non-IID}, in the case of most complex settings (NIID-1), it could be observed clearly that the proposed approach, DL-MH, reasonably outperformed the other SOTA approaches. The proposed approach performs almost similar to the upper-bound approach i.e. \(globel\_g_(t)\) \((globel_\text{labelled})\). As discussed earlier, all approaches, in NIID-2, performed better than the NIID-1 scenario. Moreover, it can also be observed that in some cases, DL-SH is slightly performing better than the DL-MH approach however, in return, it is greatly reducing the communication and computation costs. Thus it can be considered as a trade-off between accuracy and communication/computation cost.
\subsubsection{Communication overhead comparison}
Here, this paper calculates and compares the communication overhead of proposed approaches along with standard federated learning and standard distillation approaches. Here supposes the scenario of NIID-1 where each client has only 2 classes of data so in such cases the communication cost of different approaches would be as follows.

\begin{figure}[H]
    \centering
    \includegraphics[width=1\linewidth]{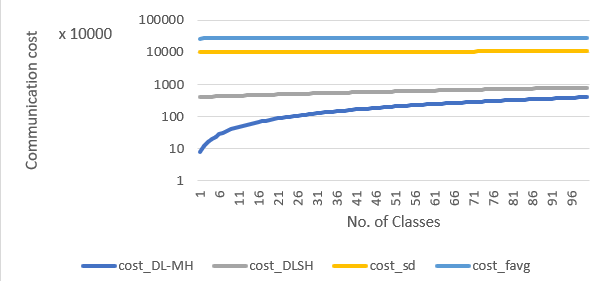}
    \caption{Comparison of communication overhead}
    \label{fig:comparison of coh}
\end{figure}

By calculating the model parameters of ResNet18 and setting the minimum communication rounds for Fedavg = 10 (though, in real scenarios including in literature, there could be hundreds of rounds), and setting client M=1. By using the Eq. \ref{eq:3.8}
\begin{align}
\text{cost}_{\text{fedavg}}= (9146954 + 9146954) \times 10 = 1.83E+08 \times w
\end{align}
By using Eq. \ref{eq:3.9} and setting the value of \(X_{\text{dist}}\)= 40000, client soft logits \(\theta=10\), weight matrix w=40000
\begin{align}
   \text{ Comm\_cost}_{\text{(DL-SH)}}=((40000*10)+40000)= 4.40E+05
\end{align}
Now by using Eq. \ref{eq:3.10} and setting the value of \(X_{\text{dist}}= 40000\), in case of non-IID, client soft logits \(\theta=2\), weight matrix w=40000, mask\_dict = 2,
\begin{align}
   \text{ Comm\_cost}_{\text{(DL-MH)}}=((40000*2)+40000+2)= 120002)
\end{align}
Here, it can be observed clearly that DL-MH greatly reduces the communication cost, particularly in non-IID scenarios and it is intuitive that in real FL scenarios, most devices would have non-IID data distribution. Moreover, this calculation is only for a single client, however, in practical scenarios, there could be hundreds of client devices. Thus, multiplying these equations with the number of clients can further exponentially reduce the communication cost as compared to standard Fedavg and standard distillation. Figure \ref{fig:comparison of coh} represents the comparison of different approaches for 100 classes against only one client. Here it can be seen that decreasing the number of classes doesn’t affect the communication cost of standard Fedavg and standard distillation however communication cost of DL-MH greatly depends upon the number of classes contained by each client.
\subsection{I-DL-MH}

\begin{figure}[H]
    \centering
    \includegraphics[width=1\linewidth]{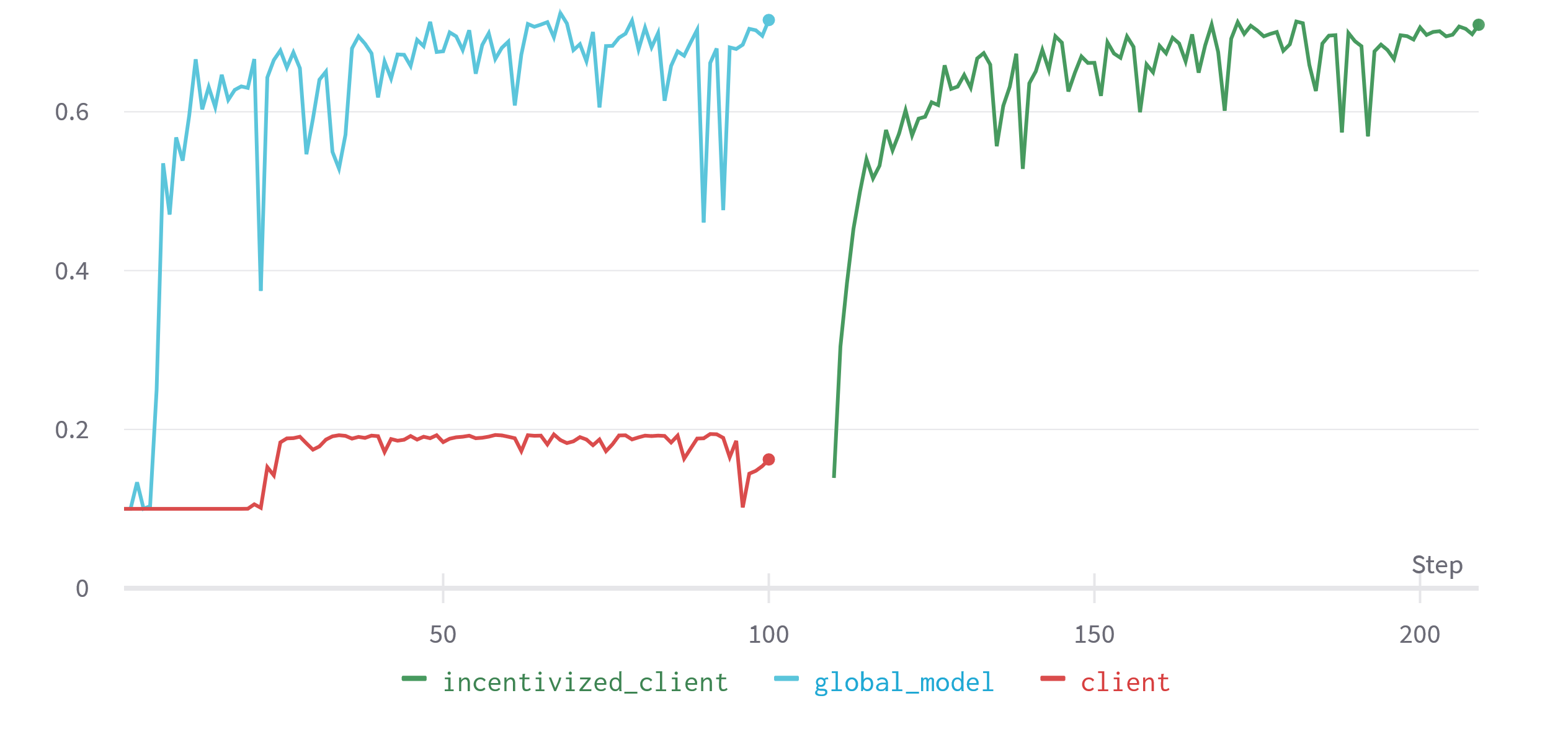}
    \caption{Test accuracy on CIFAR10 NIID-1 data distribution with ResNet18 client}
    \label{fig:4.18CIFAR10 non-IID data resnet18}
\end{figure}

This section shows the experimental results and discusses the impact of the incentive-based approach on client model performance. 
Figure \ref{fig:4.18CIFAR10 non-IID data resnet18} illustrates the effectiveness of the I-DL-MH approach where the client gets a huge performance gain after getting incentives from the global model. As discussed earlier that I-DL-MH put almost negligible communication overhead on the client devices. Here client having ResNet18 model architecture was trained on the CIFAR10 dataset with NIID-1 data distribution. It is intuitive that the client would have a maximum of 20\% test accuracy as it has samples of only two classes in NIID-1 settings. However, after applying one round of distillation from the server, the client gets a very huge performance gain
\begin{figure}[H]
    \centering
    \includegraphics[width=1\linewidth]{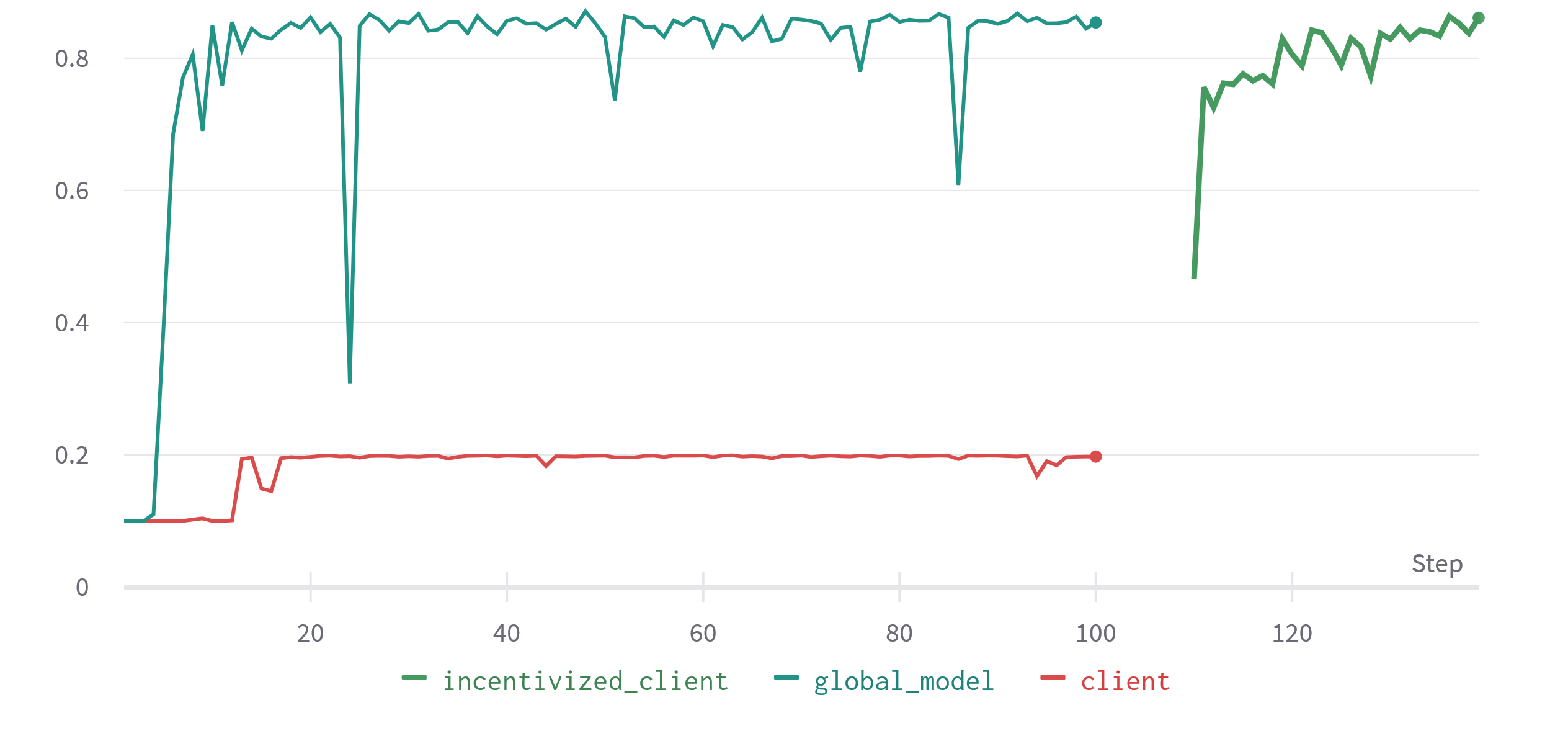}
    \caption{Test accuracy on FMNIST NIID-1 data distribution with ResNet18 client}
    \label{fig:4.19FMNIST non-IID renet18}
\end{figure}
Similarly, Figure \ref{fig:4.19FMNIST non-IID renet18} also demonstrate the effectiveness of I-DL-MH. Here it can also be observed that the client model almost gets equal to or greater than the global model accuracy with only one round of distillation from the global model and with very negligible cost.
Here let’s extend the Eq. 4.3 to calculate the additional communication cost for a client. Here w=0 because now there is no need to share the confidence matrix. \(mask_\text{dict}=0\) because all the mapping and masking procedures are performed by the server. Thus the equation becomes 

\(Comm\_cost_\text{(I-DL-MH)}= \theta\)

It means clients only download the logits from the server to distil the updated knowledge from the global model. It can be clearly seen that I-DL-MH very effectively enables the clients to get some incentives in the form of updated knowledge from the server with very negligible communication cost.

\section{Conclusion}
This research addressed three of the most fundamental challenges in federated learning (FL): statistical heterogeneity, full model heterogeneity, and the lack of effective incentive mechanisms for client participation. To this end, we proposed three complementary methodologies: DL-SH, DL-MH, and I-DL-MH, each tackling a core limitation of existing FL paradigms.

First, DL-SH introduced a confidence-matrix–based approach using binary classifiers on unlabeled public data to effectively mitigate statistical heterogeneity. This design not only improved robustness against non-IID data distributions but also incorporated one-shot communication to reduce both computation and communication costs, enabling deployment on resource-constrained devices. Empirical results demonstrated up to a 153

Second, DL-MH extended this idea to enable fully heterogeneous client models to collaborate without requiring architectural homogeneity. By introducing cost-effective mapping and masking schemes, DL-MH allowed diverse local models to participate seamlessly in global training while maintaining statistical robustness. This approach achieved a 99

Finally, I-DL-MH addressed the incentive gap in federated learning by providing participating clients with meaningful rewards in the form of distilled global knowledge, efficiently transferred through a server-side mapping and masking process. This mechanism allowed client models to achieve performance levels comparable to the global model with only a single round of communication, offering a powerful and low-cost motivation for participation. In extensive evaluations across diverse architectures (ResNet18, DenseNet, ResNet8), datasets (MNIST, FMNIST, CIFAR10, CIFAR100, CINIC10), and distributional scenarios (IID and multiple levels of non-IID), the proposed approaches consistently outperformed state-of-the-art baselines, demonstrating both scalability and generalizability.

While these contributions represent significant progress, certain limitations remain. Future research should focus on integrating privacy-preserving techniques such as differential privacy or homomorphic encryption into the proposed frameworks, exploring domain adaptation strategies to better leverage public datasets, and developing mechanisms that further reduce communication overhead in real-world deployments. Additionally, improving the deployability of the proposed methods in large-scale, dynamic FL environments remains an important open challenge.

In summary, this work advances the state of federated learning by providing a unified framework that simultaneously addresses statistical heterogeneity, model heterogeneity, and incentive design, achieving substantial gains in accuracy, efficiency, and scalability. The proposed methodologies lay a strong foundation for future FL systems that are not only more robust and efficient but also more practical and attractive for real-world adoption.

\section*{ACKNOWLEDGMENT}
The Researchers would like to thank the Deanship of Graduate Studies and Scientific Research at Qassim University for financial support (QU-APC-2025-9/1)


\bibliographystyle{unsrt}
\bibliography{library-comp}

\vfill\pagebreak

\end{document}